%% file: acl_latex.tex
\documentclass[11pt]{article}

\usepackage[final]{acl}

\usepackage{times}
\usepackage{latexsym}

\usepackage[T1]{fontenc}

\usepackage[utf8]{inputenc}

\usepackage{microtype}

\usepackage{inconsolata}

\usepackage{enumitem}
\usepackage{graphicx}
\usepackage{amsmath}
\usepackage{amsfonts}
\usepackage{makecell}

\usepackage{hyperref}       
\usepackage{url}            
\usepackage{booktabs}       
\usepackage{nicefrac}       
\usepackage{xcolor}         
\usepackage{amssymb}
\usepackage{mathtools}
\usepackage{amsthm}
\usepackage{multirow}
\usepackage{colortbl}
\usepackage{arydshln}
\usepackage{wrapfig}
\usepackage{algorithm}
\usepackage{algpseudocode}
\usepackage{float}
\usepackage{adjustbox}

\usepackage[capitalize,noabbrev]{cleveref}
\usepackage{bbding}
\usepackage{pifont}
\usepackage{minitoc}
\usepackage{booktabs}
\usepackage{minitoc}

\usepackage{bbm}
\usepackage{subfig}

\definecolor{myblue}{RGB}{224,247,250}
\definecolor{myblue2}{RGB}{186,230,251}
\definecolor{mintleaf}{RGB}{0, 184, 148}

\definecolor{new_pink}{RGB}{219, 112, 147}
\definecolor{backblue}{RGB}{210, 230, 250}

\definecolor{backred}{RGB}{255, 223, 223}

\definecolor{backgreen}{RGB}{220,244,229}

\definecolor{back_deepblue}{RGB}{180, 210, 240}

\definecolor{back_deepred}{RGB}{255, 200, 200}

\definecolor{back_deepgreen}{RGB}{190, 230, 210}

\definecolor{mygray}{gray}{0.95}

\definecolor{greentable3}{rgb}{0,0.5,0}

\definecolor{mgreen}{RGB}{6,128,67}
\definecolor{mgray}{RGB}{128,128,128}
\definecolor{mygreen}{RGB}{233,247,234}
\title{Atlas: Orchestrating Heterogeneous Models and Tools for Multi-Domain Complex Reasoning}

\author{
 \textbf{Jinyang Wu\textsuperscript{1}\thanks{\quad Equal Contribution.}\thanks{\quad Project Lead.}},
 \textbf{Guocheng Zhai\textsuperscript{1}\footnotemark[1]},
 \textbf{Ruihan Jin\textsuperscript{1}\footnotemark[1]},
 \textbf{Jiahao Yuan\textsuperscript{3}},
 \textbf{Yuhao Shen\textsuperscript{2}},\\
 \textbf{Shuai Zhang\textsuperscript{1}},
 \textbf{Zhengqi Wen\textsuperscript{1}},
 \textbf{Jianhua Tao\textsuperscript{1}\thanks{\quad Corresponding Authors.}}
\\
 \textsuperscript{1}Tsinghua University, 
 \textsuperscript{2}Zhejiang University,\\
 \textsuperscript{3}East China Normal University
\\
\texttt{wu-jy23@mails.tsinghua.edu.cn}
}

\begin{document}
\maketitle

\input{section/0_abs}

\input{section/1_intro}

\input{section/2_related_works}

\input{section/3_method}

\input{section/4_experiments}

\input{section/5_conclusion}

\input{section/limitations}

\bibliography{custom}

\input{section/6_appendix}

\end{document}

%% file: section/0_abs.tex
\begin{abstract}
The integration of large language models (LLMs) with external tools has significantly expanded the capabilities of AI agents. However, as the diversity of both LLMs and tools increases, selecting the optimal model-tool combination becomes a high-dimensional optimization challenge. Existing approaches often rely on a single model or fixed tool-calling logic, failing to exploit the performance variations across heterogeneous model-tool pairs. In this paper, we present \textbf{\textsc{Atlas}} (\textbf{A}daptive \textbf{T}ool-\textbf{L}LM \textbf{A}lignment and \textbf{S}ynergistic Invocation), a dual-path framework for dynamic tool usage in cross-domain complex reasoning. \textsc{Atlas} operates via a dual-path approach: (1) \textbf{training-free cluster-based routing} that exploits empirical priors for domain-specific alignment, and (2) \textbf{RL-based multi-step routing} that explores autonomous trajectories for out-of-distribution generalization. Extensive experiments across 15 benchmarks demonstrate that our method outperforms closed-source models like GPT-4o, surpassing existing routing methods on both in-distribution (+10.1\%) and out-of-distribution  (+13.1\%) tasks. Furthermore, our framework shows significant gains in visual reasoning by orchestrating specialized multi-modal tools.
\end{abstract}

%% file: section/1_intro.tex
\begin{figure}[t]
  \centering
  \includegraphics[width=0.48\textwidth]{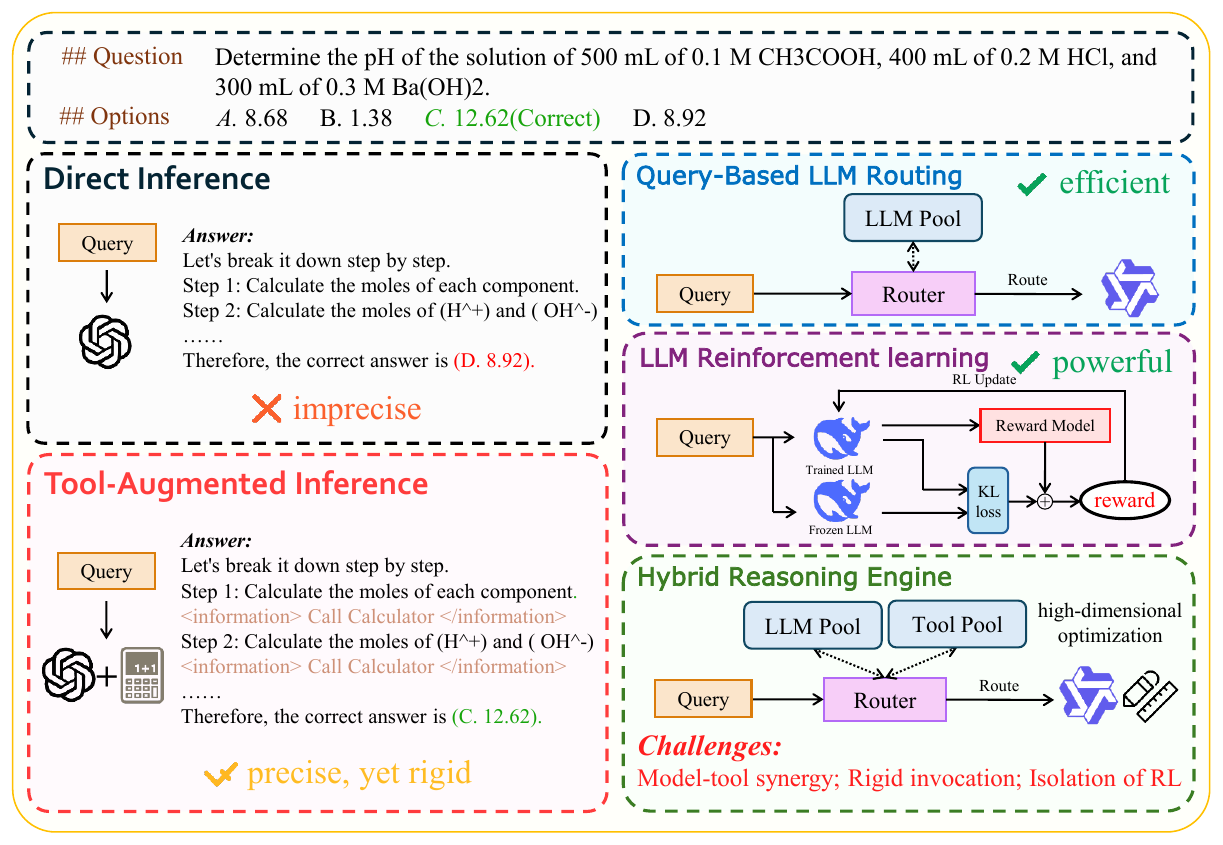}
  \caption{\textbf{Comparison of different LLM inference paradigms.} While routing (efficiency) and RL (performance optimization) present a promising approach, dynamic tool usage still faces significant challenges.}
  \label{fig:intro}
  \vspace{-0.5cm}
\end{figure}

\section{Introduction}\label{sec:intro}

Large language models (LLMs) have evolved from static problem solvers into collaborative reasoning engines through adaptive integration with external tools. These tools range from symbolic reasoning modules~\cite{feng2025retool} to real-time information retrieval APIs~\cite{ma2025tool}, significantly extending LLMs' operational capabilities. As this LLM-tool ecosystem evolves, the synergy from multiple candidates increasingly surpasses the potential of either routing in model swarms~\cite{yue2025masrouter} or tool augmentation~\cite{dong2025tool} alone, highlighting the critical need for identifying the optimal model-tool combination.

\begin{figure*}[t]
  \centering
  \includegraphics[width=0.98\textwidth]{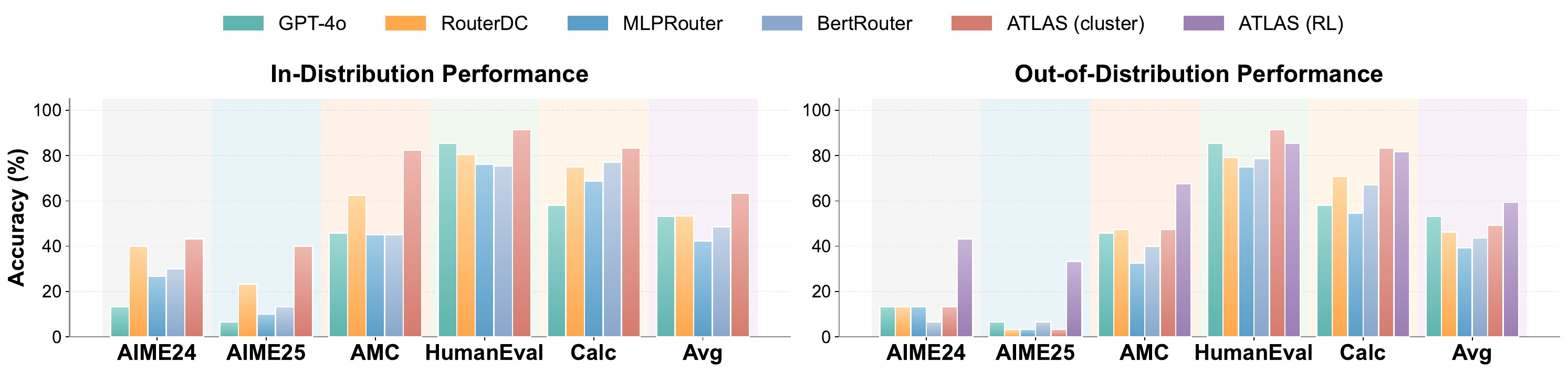}
    \caption{\textbf{Performance comparison on in-distribution and out-of-distribution settings.} Our \textsc{Atlas} method consistently outperforms all baselines across diverse datasets, demonstrating superior generalization capability.}
  \label{fig:intro_performance}
  \vspace{-0.4cm}
\end{figure*}

Recent advances have focused on different aspects of these reasoning engines separately. \textit{\textbf{For tool usage}}, existing frameworks~\cite{kong2024tptu, wu2024avatar} improve performance through task planning, yet relying on fixed logic that cannot dynamically adapt to different model capabilities or task requirements. \textit{\textbf{For LLM routing}}, methods like ZOOTER~\cite{lu2024routing} and RouterDC~\cite{chen2024routerdc} optimize model selection through reward-guided learning and dual contrastive learning. Likewise, frameworks such as HybridLLM \cite{ding2024hybrid} and RouteLLM~\cite{ong2024routellm} combine strong and weak models for cost efficiency. However, these routing methods treat models as isolated execution units and fail to incorporate external tools, which could significantly enhance task performance. \textit{\textbf{For reinforcement learning (RL)}}, methods such as RLHF~\cite{ouyang2022training} and PPO~\cite{schulman2017proximal} are explored to optimize reasoning capabilities in LLMs. RLAIF~\cite{lee2023rlaif} and DPO~\cite{rafailov2023direct} bypass explicit reward modeling, streamlining preference learning. Additionally, Router-R1~\cite{zhang2025router} allows models to deliberate internally before invoking auxiliary models. Recent works~\cite{chen2024learning, jin2025search, feng2025retool} apply RL to tool usage, but miss the opportunity to integrate both models and tools to fully harness their combined strengths.

As shown in Figure~\ref{fig:intro}, existing methods neglect the dynamic interplay of tool usage, LLM routing and RL, thus falling short especially when faced with the emerging diversity of LLMs and tools. This fundamental limitation manifests in three key challenges: (1) \textbf{Failure to leverage model-tool synergies:} LLM routing methods focus solely on model selection without integrating external tools, limiting their potential to enhance task performance; (2) \textbf{Rigid invocation and limited flexibility:} Existing tool usage methods rely on fixed, pre-configured invocation logic that hinders adaptability and scalability, preventing reasoning engines from dynamically optimizing model-tool combinations in open-domain tasks; (3) \textbf{Isolated optimization of RL:} Even advanced RL approaches focus on optimizing individual components in isolation, missing opportunities to jointly leverage model-tool synergies for complex reasoning.

To address these challenges, we propose \textbf{\textsc{Atlas}} (\textbf{A}daptive \textbf{T}ool-\textbf{L}LM \textbf{A}lignment and \textbf{S}ynergistic Invocation), a generalizable framework that dynamically orchestrates optimal model-tool combinations. Our approach employs a dual-path approach to bridge the gap between empirical knowledge and open-domain reasoning. We firstly introduce \textbf{training-free cluster-based routing} that efficiently selects model-tool pairs by leveraging domain-specific expertise within a semantic embedding space. This approach exploits historical performance patterns for rapid, accurate routing in familiar domains. For generalized scenarios where explicit priors are absent, we utilize \textbf{RL-based multi-step routing} that iteratively explores the model-tool combinations for superior execution paths. This bifurcated design effectively resolves the scalability challenges in high-dimensional search spaces while ensuring robustness. We conduct experiments on 15 benchmarks to evaluate the proposed \textsc{Atlas} in both in-distribution and out-of-distribution settings. Empirical results shown in Figure~\ref{fig:intro_performance} reveal that \textsc{Atlas} achieves a superior performance across diverse tasks, which demonstrates its effectiveness as a new paradigm for tool-augmented reasoning agents. Our primary contributions are as follows:

\begin{itemize}[leftmargin=1.18em]
    \item We introduce \textbf{\textsc{Atlas}}, a generalizable agentic framework that explicitly optimizes heterogeneous synergies between diverse LLMs and tools, enabling dynamic and adaptive tool invocation for complex reasoning tasks.
    \item We propose a dual-path design that handles both domain-specific and open-domain tasks: (1) \textbf{training-free cluster-based routing} for efficient selection using domain expertise, and (2) \textbf{RL-driven multi-step routing} for generalizing across unfamiliar tasks via iterative exploration.
    \item Experiments across 9 tasks and 15 benchmarks show that \textsc{Atlas} outperforms top-performing closed-source LLMs and powerful routing methods on multi-domain tasks and exhibits robust adaptability in multi-modal scenarios.
\end{itemize}

%% file: section/2_related_works.tex
\begin{figure*}[t]
  \centering
  \includegraphics[width=0.98\textwidth]{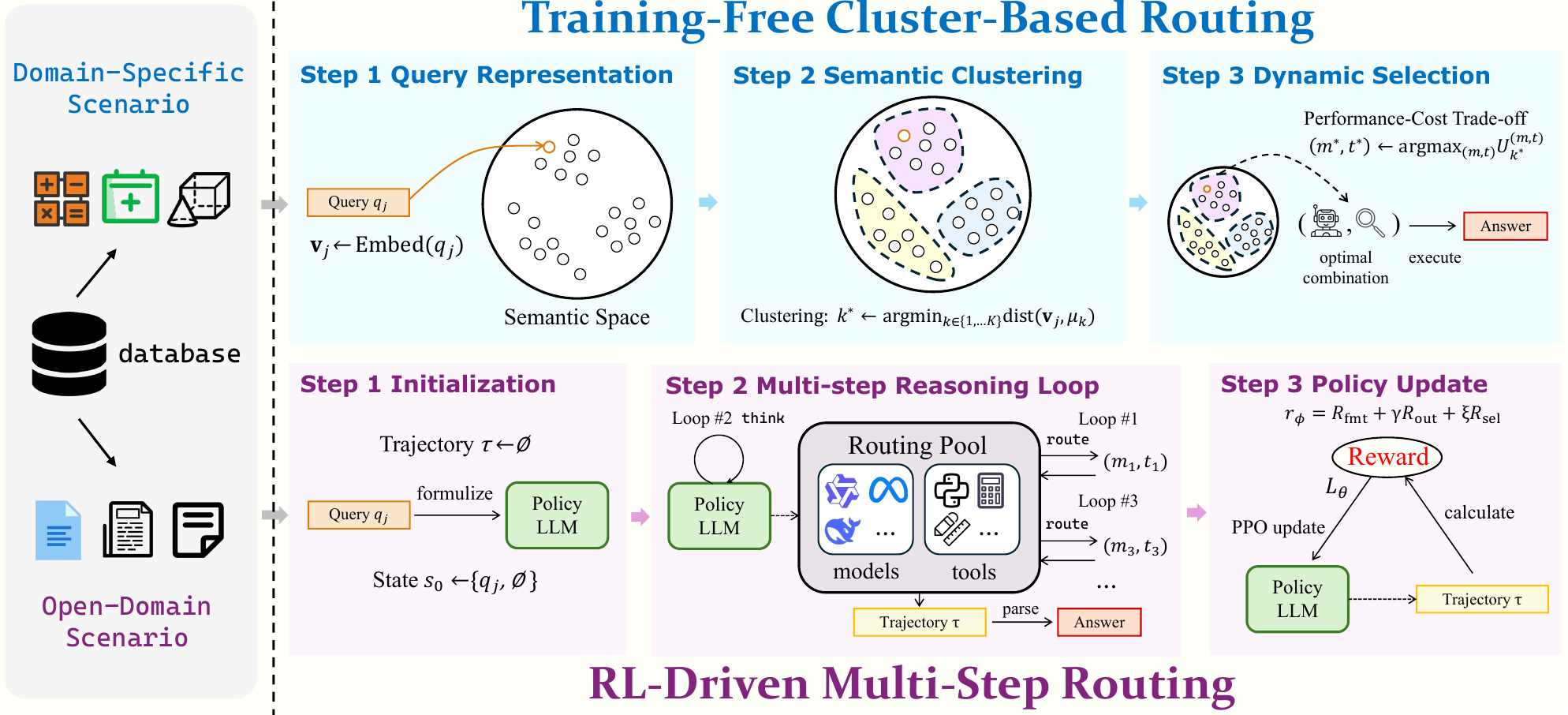}
  \caption{Overview of \textbf{A}daptive \textbf{T}ool-\textbf{L}LM \textbf{A}lignment and \textbf{S}ynergistic Invocation (\textsc{Atlas}). The framework operates via a dual-path approach: (1) Training-free Cluster-based Routing; and (2) RL-driven Multi-step Routing.}
  \label{fig:method}
  \vspace{-0.3cm}
\end{figure*}

\section{Related Work}
\label{sec:related_work}

\paragraph{Query-based LLM Routing.}
As the landscape of LLMs continues to evolve, query-based routing has become crucial in reasoning engines for balancing performance and computational efficiency by dynamically selecting the most appropriate model for each query. Early approaches rely on reward-guided \cite{lu2024routing} and contrastive learning strategies \cite{chen2024routerdc} to improve routing accuracy. Existing methods balance computational cost with performance through query-level orchestration~\cite{ding2024hybrid, ong2024routellm,zhang2025avengers}, model cascading~\cite{chen2023frugalgpt}, adaptive selection~\cite{feng2025graphrouter,wang2025icl,jin2025radial}, and budget allocation~\cite{mei2025omnirouter}. Further, the integration of routing within reasoning frameworks~\cite{yue2025masrouter, pan2025route} enhances the performance boundaries. However, these existing methods often treat LLMs as isolated execution units, neglecting the synergies between specific model capabilities and external tool interfaces. Our approach addresses this gap by jointly optimizing model-tool combinations, enabling a more adaptive, scalable, and effective reasoning engine capable of dynamically integrating the strengths of both models and tools.

\vspace{-0.1in}
\paragraph{Reinforcement Learning for LLM.}
Reinforcement learning (RL) has been widely applied to optimize LLMs for aligning with complex human preferences and improving reasoning tasks. The paradigm has evolved from reward-model-based approaches like RLHF~\cite{ouyang2022training} and PPO~\cite{schulman2017proximal} to more efficient frameworks such as DPO~\cite{rafailov2023direct}, which bypass explicit reward modeling to streamline preference learning. RL has also been applied to optimize routing decisions, with approaches like Router-R1 \cite{zhang2025router} allowing models to deliberate internally before invoking auxiliary models. Recent works~\cite{chen2024learning, jin2025search, feng2025retool} investigate the application of RL for tool usage. While these methods reveal the potential of RL in optimizing reasoning trajectories, they primarily focus on single-model or single-tool optimization, overlooking the large potential for combined synergies. \textsc{Atlas} extends this by employing RL to jointly optimize model-tool combinations, enabling more adaptive and efficient reasoning.

%% file: section/3_method.tex
\section{Methodology}\label{sec:method}

We present the \textbf{\textsc{Atlas}} framework (Figure~\ref{fig:method}), which combines a two-tier strategy: \textit{Training-free Cluster-based Routing} (§~\ref{subsec:cluster}) to enable quick decision-making; and \textit{RL-driven Multi-step Routing} (§~\ref{subsec:rl}) to handle more complex open-domain tasks that require iterative model-tool interactions.

\subsection{Training-Free Cluster-Based Routing}\label{subsec:cluster}
We hypothesize that the optimal model-tool combination is query-dependent and exhibits semantic locality. Consequently, the empirical strategy approximates the optimal routing function $f:\mathcal{Q}\to\mathcal{S}$ by leveraging historical metadata. We define the search space as the Cartesian product $\mathcal{S}=\mathcal{M}\times\mathcal{T}$, where $\mathcal{M} = \{m_1,\dots,m_M\}$ denotes the set of candidate LLMs and $\mathcal{T} = \{t_1,\dots,t_T\}$ represents the available tools.

Given $\mathcal{Q}_{\mathrm{train}} = \{q_i\}_{i=1}^N$ denote the training queries set, we map each query $q_i$ into a $D$-dimensional latent manifold using a pre-trained encoder: $\mathbf{v}_i=\mathcal{E}(q_i)\in\mathbb{R}^D$. To capture the semantic task distribution, we partition the embedding space into $K$ disjoint clusters $\{\mathcal{C}_k\}_{k=1}^K$ by minimizing the inertia:
\begin{equation}
\min_{\{\mu_k\}_{k=1}^K} \sum_{k=1}^K \sum_{\mathbf{v}_i \in \mathcal{C}_k} \| \mathbf{v}_i - \mu_k \|^2,
\end{equation}
where $\mu_k$ represents the semantic centroid of cluster $\mathcal{C}_k$. The clustering process effectively groups queries with similar reasoning requirements and tool affinities.

We derive empirical statistics from the training observations for each modal-tool pair $(m, t) \in \mathcal{S}$ within the cluster $\mathcal{C}_k$. The empirical accuracy is defined as the success rate of the pair $(m, t)$ on the cluster $\mathcal{C}_k$:
\begin{equation}
\widehat{\mathrm{Acc}}_k^{(m,t)} = \frac{1}{|\mathcal{C}_k|}\sum_{q_i\in\mathcal{C}_k} \mathbbm{1} \big[(m,t) \text{ solves } q_i\big],
\end{equation}
where $\mathbbm{1}[\cdot]$ denotes the indicator function. 

Simultaneously, we model the operational cost to account for resource consumption, which is computed based on the average token throughput observed during the profiling phase:
\begin{equation}
\widehat{\mathrm{Cost}}_k^{(m,t)}=\bar{N}_\mathrm{in}^{(m,t)} \cdot P_\mathrm{in}^{(m,t)}+\bar{N}_\mathrm{out}^{(m,t)} \cdot P_\mathrm{out}^{(m,t)},
\end{equation}
where $\bar{N}_{\mathrm{in}}$ and $\bar{N}_{\mathrm{out}}$ represent the mean input and output token counts for the cluster, while $P_{\mathrm{in}}$ and $P_{\mathrm{out}}$ denote their respective unit prices.

To facilitate a flexible trade-off between reasoning performance and inference cost, we define a cluster-level utility score $\mathcal{U}_k(m, t)$ as:
\begin{equation}
\mathcal{U}_k(m, t)=(1-\alpha)\cdot\widehat{\mathrm{Acc}}_k^{(m,t)}-\alpha\cdot\widehat{\mathrm{Cost}}_k^{(m,t)},\label{eq_acc_cost}
\end{equation}
where $\alpha\in [0,1]$ is a hyperparameter that balances the performance-cost trade-off.

At inference time, the framework performs low-latency orchestration by projecting a novel query $q_j$ into the latent manifold $\mathbf{v}_j = \mathcal{E}(q_j)$. The routing is executed via a proximal cluster lookup, where the query is assigned to $k^*=\arg\min_{k}\|\mathbf{v}_j-\mu_k\|$. Subsequently, the system retrieves the optimal model--tool pair for execution:
\begin{equation}
(m^*, t^*)=\arg\max_{(m,t)\in\mathcal{S}} \mathcal{U}_{k^*}(m, t).
\end{equation}

By caching heterogeneous synergies within the embedding space, this empirical strategy enables real-time, cost-aware tool invocation with constant-time complexity relative to the number of clusters.

\subsection{RL-Driven Multi-Step Routing}\label{subsec:rl}
While the empirical strategy excels in low-latency routing, it is inherently limited by its reliance on a single-shot decision. To address complex tasks that demand multi-round reasoning and iterative model-tool interactions, we introduce an RL-driven strategy that instantiates the router as an autonomous agent capable of interleaving internal reasoning with external invocation.

We model this process as a sequential decision task over a maximum horizon $T_\text{max}$. For a given query $q_j$, the agent maintains an evolving state $s_t = \{q_j, C_t\}$, where $C_t$ represents the accumulated context of previous reasoning trajectories and tool outputs. At each step $t$, the policy $\pi_\theta$ samples an action $a_t$ from the augmented action space $\mathcal{A}$, comprising two types of operations: (1) \textit{Internal Reasoning} (\texttt{think}), where the agent performs local chain-of-thought processing to decompose complex queries or synthesize intermediate results; and (2) \textit{Dynamic Routing} (\texttt{route}), where the agent selects a specific model-tool pair $(m, t) \in \mathcal{S}$ from the routing pool to gather external observations $o_t$. This iterative loop ensures that the agent can adaptively refine its search space based on real-time feedback from the environment until an answer is extracted or the maximum step limit is reached.

To optimize this decision-making process, we train the policy $\pi$ using Proximal Policy Optimization (PPO)~\cite{schulman2017proximal}, which maximizes the following regularized objective:
\begin{equation}
\max_{\pi}\mathbb{E}_{q\sim\mathcal{D},\tau\sim\pi} \left[r_\phi(q,\tau)-\beta\log\frac{\pi(\tau|q; \mathcal{P})}{\pi_{\text{ref}}(\tau|q; \mathcal{P})} \right],
\end{equation}
where $\tau$ is the interaction trajectory, $\pi_{\text{ref}}$ is a reference policy to ensure training stability, and $\beta$ is the KL-regularization coefficient.

We design the reward function $r_\phi$ as a composite of three finely-tuned rule-based signals (detailed in Appendix \ref{sec:appendix_reward_signals}) including format reward ($\mathcal{R}_{\text{fmt}}$), outcome reward ($\mathcal{R}_{\text{out}}$) and model selection reward ($\mathcal{R}_{\text{sel}}$), bridging the gap between structured execution and task correctness, formally: 

\begin{itemize}
    \item \textbf{Format Reward ($\mathcal{R}_{\text{fmt}}$):} A signal enforces structural integrity by penalizing trajectories that deviate from the predefined format and tool-invocation syntax.
    \item \textbf{Outcome Reward ($\mathcal{R}_{\text{out}}$):} A binary signal that directly aligns the policy with task correctness.
    \item \textbf{Model Selection Reward ($\mathcal{R}_{\text{sel}}$):} A penalty-based signal guides the agent toward optimal efficiency by penalizing the selection of sub-optimal models.
\end{itemize}

The final reward is computed as:
\begin{equation}
r_\phi = \mathcal{R}_\text{fmt}+\gamma\mathcal{R}_\text{out}+\xi\mathcal{R}_\text{sel},
\end{equation}
where $\gamma$ and $\xi$ are hyperparameters. This framework facilitates autonomous orchestration, as the model learns to assess the sufficiency of its internal state before invoking external resources. By decoupling routing logic via the $\mathcal{R}_\text{sel}$ signal, \textsc{Atlas} internalizes the fundamental alignment between domains and tool utilization rather than memorizing rigid model-tool mappings. This design ensures that the routing policy captures the essential characteristics of expertise distribution, remaining robust and generalizable even as the available tools and models evolve in dynamic environments.

%% file: section/4_experiments.tex
\input{section/4_main_results}

\section{Experiments}\label{sec:experiments}
This section presents a comprehensive evaluation of \textsc{Atlas}, covering main results across multi-domain benchmarks (§~\ref{subsec:main_results}), multi-modal visual reasoning (§~\ref{subsec:multi_modal}), model-tool pool extensions (§~\ref{subsec:dynamic_synergy}), and further analysis on reasoning boundaries, model-tool alignment preferences, and RL convergence dynamics (§~\ref{subsec:discussion}).

\subsection{Experimental Settings}\label{subsec:settings}

\paragraph{Models Selection.} 
To evaluate \textsc{Atlas}'s generalization across model architectures and scales, we select six heterogeneous open-source LLMs: Qwen2.5-7B-Instruct~\cite{yang2024qwen2}, Llama-3.1-8B-Instruct~\cite{dubey2024llama}, InternLM3-8B-Instruct~\cite{cai2024internlm2}, DeepSeek-R1-Distill-Qwen-7B~\cite{guo2025deepseek}, Qwen2.5-Coder-7B-Instruct~\cite{hui2024qwen2}, and a multi-modal LLM Qwen3-8B-VL-Instruct~\cite{yang2025qwen3}. This diverse selection allows us to observe how different models synergize with specific external tools.

\paragraph{Tool Definition.} 
We introduce two tool sets for textual and visual reasoning:

\begin{itemize}
    \item \textbf{Foundation Tools}: This set includes four essential tools: (1) \textit{Code Interpreter}, a Python execution environment for algorithmic and logical verification; (2) \textit{Web Search} for retrieving real-time open-domain information; (3) \textit{Calculator} for high-precision numerical computation; and (4) \textit{Process Reward Model (PRM)} for scoring and ranking model outputs.
    \item \textbf{Multi-modal Tools}: (1) \textit{Qwen3-Chart} for chart data extraction; (2) \textit{Qwen3-Counting} for enumerating objects in images; (3) \textit{Qwen3-Geo} for parsing geometric properties and performing post-hoc self-verification of geometric proofs; and (4) \textit{Hunyuan-OCR}~\citep{team2025hunyuanocr} for text extraction from images. The first three tools use Qwen3-8B-VL with task-specific prompts, due to the underperformance of most existing specialized tools.
\end{itemize}

\paragraph{Benchmarks and Baselines.}
We evaluate on multi-domain tasks: (1) mathematical reasoning: AIME2024~\cite{AIME2024}, AIME2025~\cite{AIME2025}, AMC~\cite{lightman2023let}; (2) code generation: HumanEval~\cite{chen2021evaluating}, MBPP~\cite{austin2021program}; (3) arithmetic reasoning: Calculator~\cite{wu2025tool}; (4) commonsense reasoning: NQ~\cite{kwiatkowski2019natural}, WebQ~\cite{webq}; (5) logical reasoning: LogiQA2~\cite{liu2023logiqa}; (6) scientific reasoning: GPQA~\cite{rein2024gpqa}. Furthermore, we extend our evaluations to multi-modal benchmarks, including ChartQA~\cite{masry2022chartqa}, Geometry3K~\cite{lu2021inter}, TallyQA~\cite{acharya2019tallyqa}, CountBench~\cite{paiss2023teaching}, and TableVQA~\cite{kim2024tablevqa}. We use accuracy as the primary metric. Baselines include \textbf{Zero-shot/Few-shot} \textbf{Router}, \textbf{Random Router}, \textbf{RouterDC}~\cite{chen2024routerdc}, \textbf{MLPRouter}~\cite{hu2024routerbench}, and \textbf{BertRouter}~\cite{ong2024routellm}. Details are provided in Appendix~\ref{sec:appendix_experimental_details}.

\paragraph{Implementation Details.}
For RL experiments, we adopt Qwen2.5-3B-Instruct as the policy model for model-tool selection. The policy is optimized with a batch size of 32 for 250 training steps, and the learning rate is set to $1\times10^{-6}$. More details are provided in Appendix~\ref{subsec:appendix_implementation_details}.

\begin{table*}[h]
    \scriptsize
    \centering
    \caption{\textbf{Performance evaluation under dynamic routing pool extensions.} $\dag$ denotes results after integrating domain-specialized models (Llama-3.1-8B-UltraMedical, Qwen2.5-Math-7B-Instruct) and an Outcome Reward Model into the routing pool. $\ddag$ marks in-domain benchmarks; all others are out-of-domain. Best results are in \textbf{bold}.}
    \label{tab:extension_results}
    \resizebox{1.0\linewidth}{!}{
    \begin{tabular}{lccccccccccc}
    \toprule
    \multirow{2}{*}{\textbf{Method}} & \multicolumn{3}{c}{\textbf{Math Reasoning}} & \multicolumn{2}{c}{\textbf{Code}} & \textbf{Arith.} & \multicolumn{2}{c}{\textbf{Common.}} & \textbf{Logic} & \textbf{Sci.} & \multirow{2}{*}{\textbf{Avg.}} \\
    \cmidrule(lr){2-4} \cmidrule(lr){5-6} \cmidrule(lr){7-7} \cmidrule(lr){8-9} \cmidrule(lr){10-10} \cmidrule(lr){11-11}
    & AIME24 & AIME25 & AMC & Human. & MBPP$^\ddag$ & Calc.$^\ddag$ & NQ$^\ddag$ & WebQ & LQA2 & GPQA & \\
    \midrule
    ZS Router         & 13.3 & 6.7  & 32.5 & 53.0 & 64.2 & 55.7 & 29.2 & 39.2 & 45.3 & 24.6 & 36.4 \\
    \rowcolor[RGB]{236,244,252}
    ZS Router$^\dag$  & 20.0 & 13.3  & 37.5 & 52.4 & 63.1 & 55.0 & 28.7 & 38.9 & 45.9 & 25.7 & 38.0 \\
    \midrule
    FS Router         & 23.3 & 13.3 & 40.0 & 68.9 & 64.7 & 47.2 & 27.3 & 35.8 & 40.8 & 25.9 & 38.7 \\
    \rowcolor[RGB]{236,244,252}
    FS Router$^\dag$  & 26.7 & 16.7  & 47.5 & 70.7 & 63.8 & 46.5 & 25.9 & 36.2 & 41.7 & 25.0 & 40.0 \\
    \midrule
    RandomRouter      & 6.7 & 3.3 & 15.0 & 37.8 & 52.6 & 40.2 & 25.3 & 32.1 & 49.2 & 30.6 & 29.3 \\
    \rowcolor[RGB]{236,244,252}
    RandomRouter$^\dag$ & 3.3 & 3.3  & 17.5 & 35.4 & 52.0 & 41.3 & 22.7 & 31.5 & 49.8 & 30.1 & 28.7 \\
    \midrule
    BertRouter        & 26.7 & 16.7 & 42.5 & 76.8 & 72.6 & 62.7 & 35.4 & 49.8 & 52.5 & 33.3 & 46.9 \\
    \rowcolor[RGB]{236,244,252}
    BertRouter$^\dag$ & 33.3 & 20.0  & 50.0 & 75.0 & 73.0 & 61.3 & 36.2 & 50.1 & 53.4 & 32.4 & 48.4 \\
    \midrule
    \textsc{Atlas} (RL) & 43.3 & 33.3 & 67.5 & \textbf{85.4} & \textbf{81.8} & 81.6 & 44.1 & 52.2 & 62.7 & 42.0 & 59.4 \\
    \rowcolor[RGB]{245, 238, 248}
    \textbf{\textsc{Atlas} (RL)}$^\dag$ & \textbf{50.0} & \textbf{40.0} & \textbf{70.0} & 84.2 & \textbf{81.8} & \textbf{82.4} & \textbf{45.3} & \textbf{52.8} & \textbf{64.8} & \textbf{45.1} & \textbf{61.7} \\
    \bottomrule
    \end{tabular}}
\end{table*}

\subsection{Main Results}\label{subsec:main_results}
Table \ref{tab:main_results} presents a comprehensive evaluation of our framework against various routing baselines across in-distribution and out-of-distribution tasks.

\subsubsection{In-Distribution Performance}
Under the in-distribution setting, where training data for all tasks is accessible, \textsc{Atlas}(cluster) achieves 63.5\% average accuracy, surpassing the strongest baseline RouterDC by 10.1\%. This advantage is pronounced on rigorous mathematical reasoning: \textsc{Atlas} achieves 40.0\% on AIME25 and 82.5\% on AMC (+16.7\% and +20.0\% over RouterDC). Notably, \textsc{Atlas}(cluster) exceeds GPT-4o (53.1\%) and approaches GPT-4.1 (63.0\%), demonstrating that strategic model–tool orchestration enables a reasoning engine of smaller-scale models to rival larger proprietary systems.

This performance stems from exploiting rich empirical priors through semantic embedding. By mapping queries into structured clusters and caching historical performance patterns, the framework achieves near-optimal task-configuration alignment. In contrast, supervised routers like BertRouter and MLPRouter struggle with non-linear decision boundaries in heterogeneous model-tool spaces. Their classification-based selection fails to capture nuanced synergies from domain-specific pairings, resulting in suboptimal routing.

\subsubsection{Generalization Scenarios}
When facing out-of-distribution (OOD) challenges, \textsc{Atlas}(cluster) suffers significant degradation (e.g., dropping from 40.0\% to 3.3\% on AIME25) as well as other baselines, whereas \textsc{Atlas}(RL) maintains an average accuracy of 59.4\% with 10.2\% higher than \textsc{Atlas}(cluster) (49.2\%) and 13.1\% higher than RouterDC (46.3\%). The gap is most striking in mathematical reasoning: on AIME24 and AIME25, \textsc{Atlas}(RL) sustains 43.3\% and 33.3\% accuracy, respectively, while the clustering method achieves only 13.3\% and 3.3\% (a 10× difference). This indicates that the RL path learns transferable collaborative decision principles rather than task-specific mappings.

\textsc{Atlas}(RL) autonomously explores effective trajectories through multi-faceted reward signals, learning generalizable patterns of model-tool synergies: when to invoke symbolic tools for verification or route to reasoning-specialized models rather than memorizing task-specific mappings. This enables robust transfer, maintaining competitive performance on unfamiliar tasks like AIME24 (43.3\%) and GPQA (42.0\%), approaching or exceeding GPT-4o despite using only 7B and 8B models. These results confirm that RL-driven component provides essential generalization capability, effectively bridging established domain expertise and unseen reasoning challenges.

\subsection{Multi-modal Tool Orchestration}\label{subsec:multi_modal}
\begin{figure}[t]
  \includegraphics[width=\columnwidth]{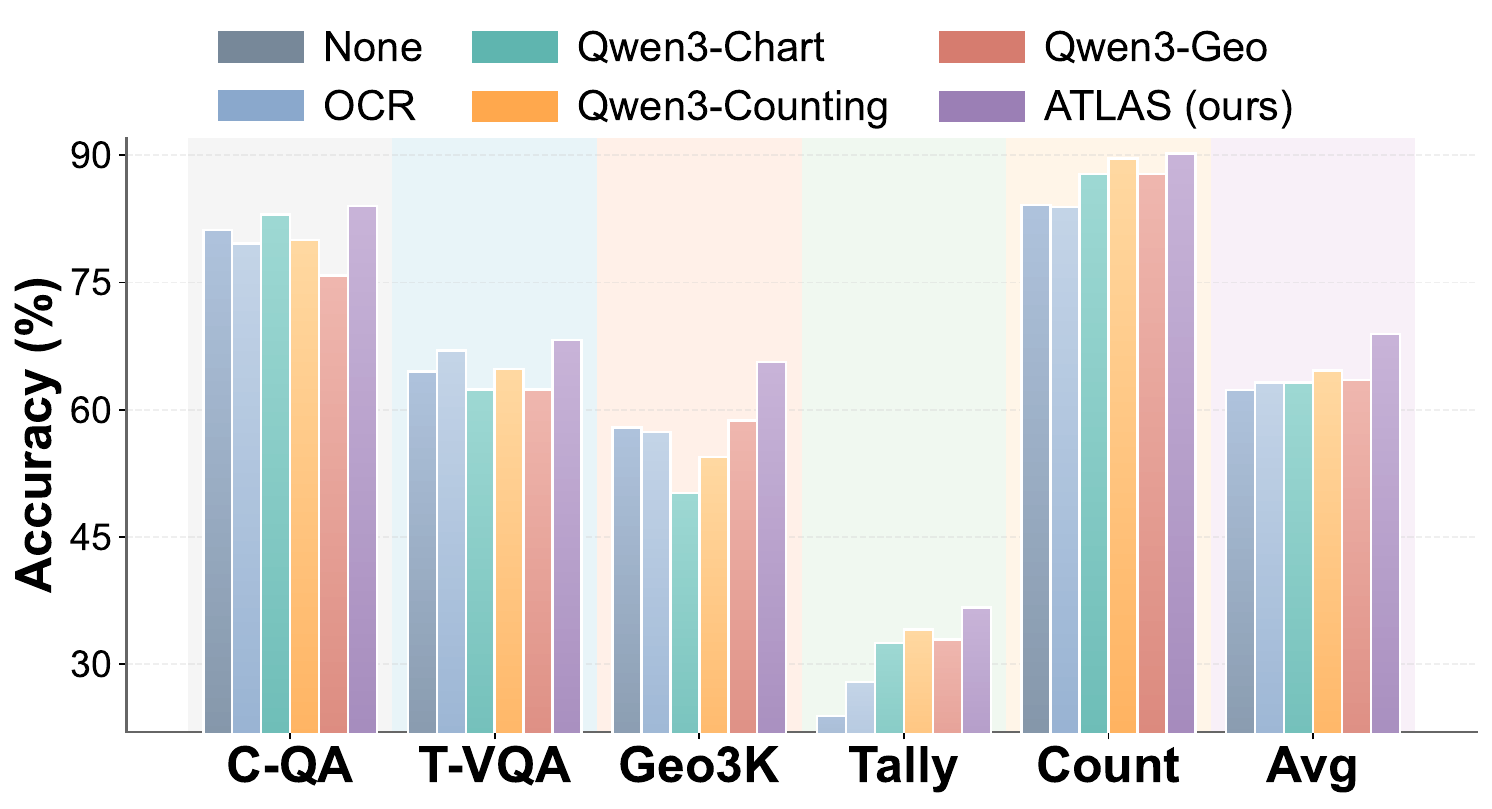}
\caption{\textbf{Performance comparison of \textsc{Atlas} against single-tool baselines across multi-modal benchmarks}. `None' denotes direct reasoning without any tools. \textsc{Atlas} achieves the highest accuracy.}
  \label{fig:multimodal}
  \vspace{-0.5cm}
\end{figure}

To evaluate \textsc{Atlas} on multi-modal tasks, we benchmark it against single-tool baselines across five visual understanding and reasoning datasets, as shown in Figure~\ref{fig:multimodal}. \textsc{Atlas} achieves an average accuracy of 68.9\% through dynamic tool invocation, outperforming the strongest single-tool baseline by 4.3\%. Notably, \textsc{Atlas} surpasses all individual tool in each task category. For example, exceeding the best single tool (Qwen3-Chart, 83.0\%) on ChartQA and overcoming the performance limitations of single tools (e.g., Qwen3-Chart only achieves 50.2\% on Geometry3K). This reveals that adaptive model-tool routing effectively integrates internal reasoning with external tool augmentation, thus establishing strong effectiveness on complex multi-modal tasks. Detailed results are provided in Appendix~\ref{appendix_multimodal}.

\subsection{Generalization Toward Dynamic Model-Tool Synergy}\label{subsec:dynamic_synergy}
A practical orchestration framework must accommodate an evolving ecosystem where new models and tools are continuously introduced. To evaluate this extensibility, we expand the routing pool with three additional components: Llama-3.1-8B-UltraMedical~\citep{zhang2024ultramedical} for biomedical reasoning, Qwen2.5-Math-7B-Instruct~\citep{yang2024qwen2math} for mathematical problem-solving, and an Outcome Reward Model for solution verification. Notably, our policy is trained exclusively on the original pool of 5 models and 4 tools; the newly added components are introduced only at inference time without any retraining. This extension substantially increases the combinatorial search space, posing a more challenging routing problem.

As shown in Table~\ref{tab:extension_results}, \textsc{Atlas}(RL) exhibits strong adaptability, improving from 59.4\% to 61.7\% (+2.3\%) after pool extension. Gains are most pronounced on mathematical benchmarks: AIME24 (+6.7\%) and AIME25 (+6.7\%), confirming effective utilization of the newly added math-specialized model and verification tool. In contrast, baseline methods show limited or inconsistent responses: BertRouter gains only +1.5\%, while RandomRouter degrades due to the expanded search space. This disparity arises because classifier-based routers learn fixed decision boundaries that become misaligned with new candidates, whereas \textsc{Atlas} learns transferable routing principles through RL exploration, enabling seamless integration of new components without retraining.

\begin{table*}[h]
    \scriptsize
    \centering
    \caption{\textbf{Reasoning capacity boundary analysis of \textsc{Atlas}(RL).} We report the pass@k metrics across diverse benchmarks to evaluate the exploration  ($k=1$) and the potential reasoning upper bound ($k=16$).}
    \label{tab:reasoning_boundary}
    \resizebox{1.0\linewidth}{!}{
    \begin{tabular}{llccccccccccc}
    \toprule
     & \multicolumn{3}{c}{\textbf{Math Reasoning}} & \multicolumn{2}{c}{\textbf{Code}} & \textbf{Arith.} & \multicolumn{2}{c}{\textbf{Common.}} & \textbf{Logic} & \textbf{Sci.} & \multirow{2}{*}{\textbf{Avg.}} \\
    \cmidrule(lr){2-4} \cmidrule(lr){5-6} \cmidrule(lr){7-7} \cmidrule(lr){8-9} \cmidrule(lr){10-10} \cmidrule(lr){11-11}
    & AIME24 & AIME25 & AMC & Human. & MBPP$^\ddag$ & Calc.$^\ddag$ & NQ$^\ddag$ & WebQ & LQA2 & GPQA & \\
    \midrule
    \rowcolor{gray!8}\multicolumn{12}{c}{\textit{Pass@1 Results with/without \textsc{Atlas} RL Training}} \\
    w/o  & 13.3 & 6.7 & 32.5 & 53.0 & 64.2 & 55.7 & 29.2 & 39.2 & 45.3 & 24.6 & 36.4 \\
    w  & \textbf{43.3} & \textbf{33.3} & \textbf{67.5} & \textbf{85.4} & \textbf{81.8} & \textbf{81.6} & \textbf{44.1} & \textbf{52.2} & \textbf{62.7} & \textbf{42.0} & \textbf{59.4} \\
    \rowcolor[RGB]{245, 238, 248}
    $\bigtriangleup$ & +30.0 & +26.6 & +35.0 & +32.4 & +17.6 & +25.9 & +14.9 & +13.0 & +17.4 & +17.4 & +23.0 \\ 
    \midrule
    \midrule
    \rowcolor{gray!8}\multicolumn{12}{c}{\textit{Pass@16 Results with/without \textsc{Atlas} RL Training}} \\
    w/o & 16.7 & 13.3 & 40.0 & 73.1 & 73.9 & 70.6 & 36.8 & 48.8 & 47.0 & 27.2 & 44.7 \\
    w & \textbf{50.0} & \textbf{36.7} & \textbf{75.0} & \textbf{89.6} & \textbf{84.5} & \textbf{83.3} & \textbf{46.9} & \textbf{54.9} & \textbf{64.4} & \textbf{45.8} & \textbf{63.1} \\
    \rowcolor[RGB]{245, 238, 248}
    $\bigtriangleup$ & +33.3 & +23.4 & +35.0 & +16.5 & +10.6 & +12.7 & +10.1 & +6.1 & +17.4 & +18.6 & +18.4 \\ 
    \bottomrule
    \end{tabular}}
    \vspace{-0.1in}
\end{table*}

\begin{figure*}[t]
    \centering
    \subfloat[Average LLM API Calls]{
    \includegraphics[width=0.32\textwidth]{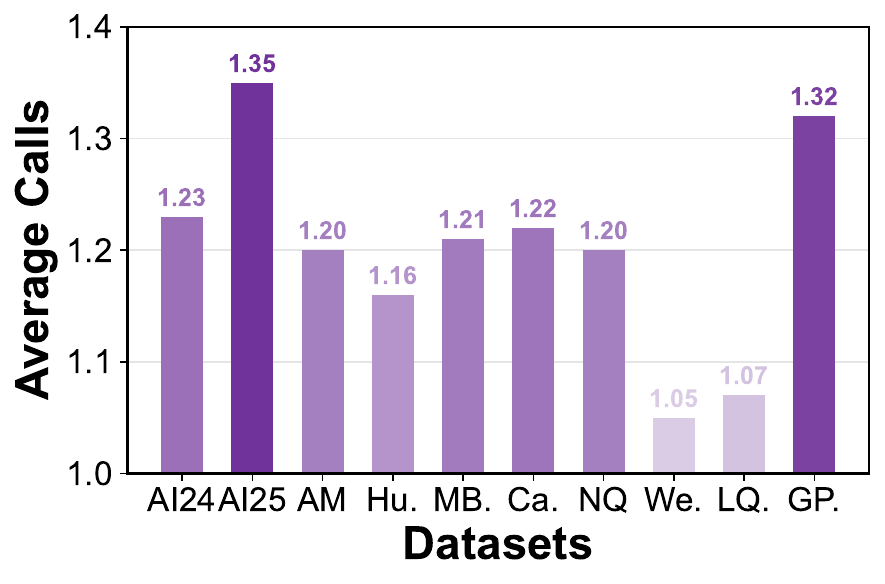}
    \label{fig:api_calls}}
    \hfill
    \subfloat[Reward Convergence]{
    \includegraphics[width=0.32\textwidth]{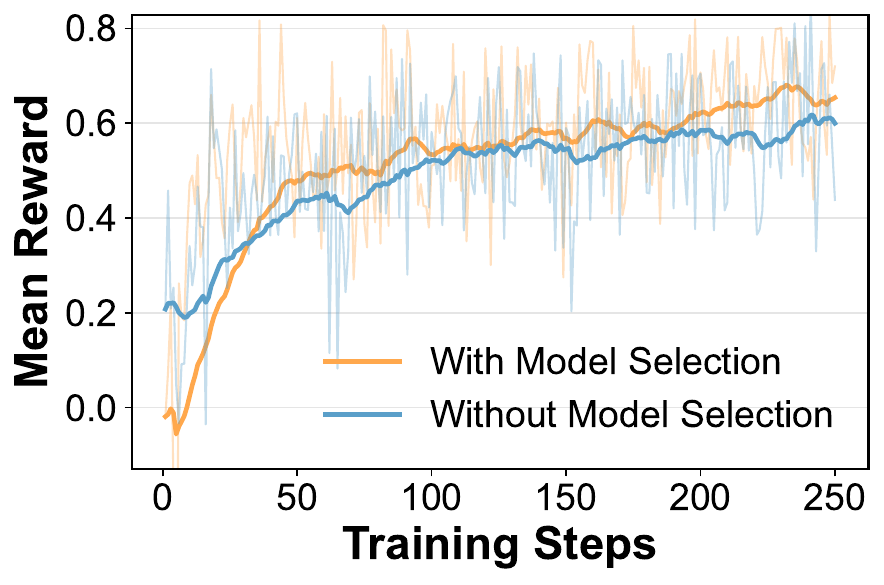}
    \label{fig:reward_conv}}
    \hfill
    \subfloat[Entropy Loss]{
    \includegraphics[width=0.32\textwidth]{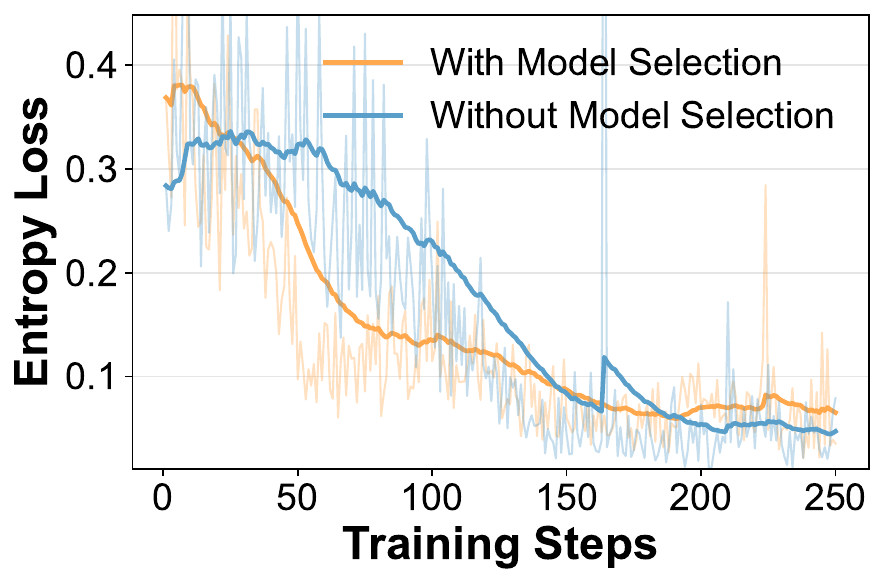}
    \label{fig:entropy_loss}}
    \caption{\textbf{Analysis of LLM API call count and \textsc{Atlas}(RL) training dynamics.}}
    \label{fig:convergence_overall}
    \vspace{-0.1in}
\end{figure*}

\subsection{Discussion}\label{subsec:discussion}
\paragraph{Evaluation on Reasoning Capacity Boundary.}
Inspired by~\cite{yue2025does}, we implement the pass@k metric to measure the reasoning capacity boundary of \textsc{Atlas}(RL), where pass@k equals 1 if at least one of k sampled outputs passes verification. As shown in Table~\ref{tab:reasoning_boundary}, RL training yields an absolute improvement of +23.0\% in pass@1 accuracy (from 36.4\% to 59.4\%), demonstrating significantly optimized exploration efficiency. At pass@16, the upper bound reaches 63.1\% (+3.7\%), indicating that \textsc{Atlas}(RL) already operates near its reasoning capacity ceiling, efficiently converging to optimal solutions without requiring extensive sampling. The trained model maintains substantial advantages across all tasks even at pass@16, with gains ranging from +6.1\% to +35.0\%, confirming that RL training effectively enhances agentic reasoning potential.

\paragraph{Analysis of LLM API Call Count.}
As illustrated in Figure~\ref{fig:api_calls}, \textsc{Atlas} exhibits highly task-adaptive invocation patterns. For challenging reasoning-intensive tasks like AIME25 and GPQA, API calls increase significantly as the RL policy allocates higher computational budgets through multi-round routing and verification. Conversely, for straightforward retrieval tasks like WebQ and NQ, call counts remain minimal. This differentiated distribution confirms that \textsc{Atlas} balances reasoning performance and inference cost, effectively suppressing redundant invocations where simpler models or fewer rounds suffice.

\paragraph{Analysis of RL Training Convergence.}\label{para:training}
We validate the RL-driven routing policy's stability through reward and entropy evolution during training. Figure~\ref{fig:reward_conv} shows that incorporating model selection reward ($\mathcal{R}_{\text{sel}}$) yields faster convergence to a higher plateau compared to the baseline, guiding the agent toward higher-yield decision regions. Figure~\ref{fig:entropy_loss} demonstrates that the \textsc{Atlas} configuration achieves a much sharper reduction in entropy compared to the ablation group, reaching a lower terminal value. This indicates that the router successfully transitions from stochastic exploration to a deterministic, high-confidence decision-making state, ensuring both the robustness and predictability of the routing process.

\paragraph{Statistical Significance.}
To verify that the observed gains are not due to chance, we conducted the Wilcoxon signed-rank test~\cite{wilcoxon1992individual}---a nonparametric paired test that requires no normality assumption---by pairing \textsc{Atlas} against the strongest baseline (RouterDC) across all benchmarks in Table~\ref{tab:main_results}. For in-distribution settings the test yields $p = 9.7 \times 10^{-4}$ ($p < \alpha{=}0.05$), decisively rejecting $H_0$ and confirming that \textsc{Atlas}'s superiority is statistically significant. Equivalent significance ($p < 0.05$) is observed in the OOD setting. Variance analysis across three repeated runs further confirms stability: e.g., \textsc{Atlas}(Cluster) achieves $82.5 \pm 2.5$ on AMC and $91.5 \pm 1.6$ on HumanEval (in-distribution).

\paragraph{More Discussion.}
Due to space, we include more discussions in Appendix, including detailed multimodal results (\ref{appendix_multimodal}), test-time scaling (\ref{appendix_tts}), analysis of model-tool preferences (\ref{appendix_invocation_preference}), ablation on reward design (\ref{appendix_ablation_study}), and sensitivity analysis (\ref{appendix_sensitivity_analysis}).

%% file: section/4_main_results.tex
\begin{table*}[t]
\small
\centering
\caption{\textbf{Performance comparison across diverse tasks and domains.} \textit{In-Distribution}: All datasets have training data available, so evaluation is in-distribution. \textit{Out-of-Distribution}: Models are trained only on Calc., NQ, and MBPP (in-distribution, marked as $\ddag$), then evaluated on all datasets (out-of-distribution for AIME24, AIME25, AMC, HumanEval, WebQ, LQA2, and GPQA). Zero-shot Router uses direct prompting without examples, while Few-shot Router uses prompting with examples. The best results are highlighted in \textbf{bold}.}
\label{tab:main_results}
\resizebox{1.0\linewidth}{!}{
\begin{tabular}{lccccccccccc}
\toprule
\multirow{2}{*}{\textbf{Method}} & \multicolumn{3}{c}{\textbf{Math Reasoning}} & \multicolumn{2}{c}{\textbf{Code}} & \textbf{Arith.} & \multicolumn{2}{c}{\textbf{Common.}} & \textbf{Logic} & \textbf{Sci.} & \multirow{2}{*}{\textbf{Avg.}} \\
\cmidrule(lr){2-4} \cmidrule(lr){5-6} \cmidrule(lr){7-7} \cmidrule(lr){8-9} \cmidrule(lr){10-10} \cmidrule(lr){11-11}
& AIME24 & AIME25 & AMC & Human. & MBPP$^\ddag$ & Calc.$^\ddag$ & NQ$^\ddag$ & WebQ & LQA2 & GPQA & \\
\midrule
\rowcolor{gray!8}\multicolumn{12}{c}{\textit{Closed-Source Models}}\\
Gemini2.5-Pro & 92.0 & 86.7 & 62.5 & 81.5 & 83.7 & 64.7 & 59.2 & 63.5 & 78.9 & 84.0 & 75.6 \\ 
GPT-5          & 93.3 & 94.6 & 97.5 & 93.4 & 98.4 & 82.9 & 59.3 & 61.5 & 83.8 & 85.7 & 85.0 \\ 
GPT-4.1        & 46.7 & 33.3 & 82.5 & 92.1 & 57.7 & 62.0 & 54.5 & 61.5 & 78.2 & 62.1 & 63.0 \\ 
GPT-4o         & 13.3 & 6.7 & 45.8 & 85.4 & 82.6 & 58.1 & 59.4 & 63.0 & 72.9 & 44.4 & 53.1 \\ 
\midrule
\midrule
\rowcolor{gray!8}\multicolumn{12}{c}{\textit{Training-free Baselines}} \\
ZS Router        & 13.3 & 6.7  & 32.5 & 53.0 & 64.2 & 55.7 & 29.2 & 39.2 & 45.3 & 24.6 & 36.4 \\
FS Router        & 23.3 & 13.3 & 40.0 & 68.9 & 64.7 & 47.2 & 27.3 & 35.8 & 40.8 & 25.9 & 38.7 \\
Random Router    & 6.7  & 3.3  & 15.0 & 37.8 & 52.6 & 40.2 & 25.3 & 32.1 & 49.2 & 30.6 & 29.3 \\
\midrule
\midrule
\rowcolor{gray!8}\multicolumn{12}{c}{\textit{In-Distribution Performance}} \\
ReAct~\citep{yao2023react}
  & 13.3   & 13.3 & 37.5 & 64.6   & 71.4 & 48.5   & 31.3   & 38.4   & 45.3 & 35.3 & 39.9 \\
Reflexion~\cite{shinn2023reflexion}
  & 13.3   & 13.3 & 40.0 & 61.0   & 73.4 & 52.1   & 34.6   & 36.8   & 46.7 & 33.9 & 40.5 \\
RouterDC         & 40.0 & 23.3 & 62.5 & 80.5 & 77.7 & 74.9 & 41.2 & 47.6 & 47.2 & 39.1 & 53.4 \\
MLPRouter        & 26.7 & 10.0 & 45.0 & 76.2 & 68.7 & 48.2 & 32.1 & 40.4 & 41.2 & 34.8 & 42.3 \\
BertRouter       & 30.0 & 13.3 & 45.0 & 75.4 & 72.1 & 77.1 & 38.9 & 50.4 & 47.1 & 36.6 & 48.6 \\
\rowcolor[RGB]{245, 238, 248}
\textbf{\textsc{Atlas} (cluster)} & \textbf{43.3} & \textbf{40.0} & \textbf{82.5} & \textbf{91.5} & \textbf{83.6} & \textbf{83.3} & \textbf{43.8} & \textbf{53.6} & \textbf{66.8} & \textbf{46.4} & \textbf{63.5} \\
\midrule
\rowcolor{gray!8}\multicolumn{12}{c}{\textit{Out-of-Distribution Performance}} \\
ReAct
  & 6.7   & 3.3  & 32.5 & 66.5   & 72.5 & 52.4   & 32.8   & 36.0   & 40.2 & 27.2 & 37.0 \\
Reflexion
  & 6.7   & 6.7  & 37.5 & 64.0   & 72.9 & 50.7   & 34.9   & 37.2   & 43.9 & 28.8 & 38.3 \\
RouterDC         & 13.3 & 3.3  & 47.5 & 79.2 & 78.7 & 70.8 & 40.1 & 50.8 & 50.4 & 28.6 & 46.3 \\
MLPRouter        & 13.3 & 3.3  & 32.5 & 75.0 & 67.7 & 54.6 & 37.3 & 43.7 & 38.9 & 26.8 & 39.3 \\
BertRouter       & 6.7  & 6.7  & 40.0 & 78.7 & 79.0 & 67.0 & 38.9 & 51.4 & 40.3 & 27.7 & 43.6 \\
\rowcolor[RGB]{236,244,252}
\textbf{\textsc{Atlas} (cluster)} & 13.3 & 3.3 & 47.5 & \textbf{91.5} & \textbf{83.6} & \textbf{83.3} & 43.8 & 51.4 & 45.6 & 29.0 & 49.2 \\
\rowcolor[RGB]{245, 238, 248}
\textbf{\textsc{Atlas} (RL)} & \textbf{43.3} & \textbf{33.3} & \textbf{67.5} & 85.4 & 81.8 & 81.6 & \textbf{44.1} & \textbf{52.2} & \textbf{62.7} & \textbf{42.0} & \textbf{59.4} \\
\bottomrule
\end{tabular}}
\end{table*}

%% file: section/5_conclusion.tex
\section{Conclusion}\label{sec:conclusion}

We present \textbf{\textsc{Atlas}}, a generalizable framework for dynamic model-tool alignment through dual-path architecture: training-free cluster-based routing for domain-specific efficiency and RL-driven exploration for open-domain adaptability. \textsc{Atlas} rivals or exceeds powerful closed-source models across diverse benchmarks, demonstrating a paradigm shift from model-centric scaling to ecosystem-centric orchestration. Experimental results show that strategic coordination of heterogeneous model-tool combinations unlocks superior reasoning while maintaining efficiency. As model and tool ecosystems continue to evolve, such orchestration reasoning systems will become essential for next-generation autonomous agents that address complex real-world challenges.

%% file: section/limitations.tex
\section*{Limitations}\label{sec:limitation}
While \textsc{Atlas} demonstrates strong performance across diverse benchmarks, several limitations warrant discussion. First, our current evaluation focuses primarily on text-based and visual reasoning tasks; extending to other modalities (e.g., audio, video) remains unexplored. Second, our framework assumes reliable API access to candidate models and tools-network latency or service unavailability in real-world deployments may impact performance. We plan to investigate more lightweight policy architectures and robust fallback mechanisms in future work.

\section*{Acknowledgments}
This work is supported by the National Natural Science Foundation of China (No. U2436210, No. 62322120).

\section*{Ethical Considerations}
All datasets, models, and tools utilized in this work are derived from publicly available resources with proper citations, involving no private or sensitive information. \textsc{Atlas} consists of two components: a training-free cluster-based router and an RL-trained policy model that learns to orchestrate existing LLMs and tools. While the policy model is trained to make routing decisions, the underlying candidate models and tools remain unmodified. Consequently, our framework inherits the potential biases, safety limitations, and ethical concerns present in these constituent components. \textsc{Atlas} itself does not introduce new harmful capabilities beyond those already existing in the routing pool. We recommend that practitioners carefully evaluate all candidate models and tools for compliance with ethical guidelines, and apply appropriate safety measures when deploying \textsc{Atlas} in real-world applications.


%% file: section/6_appendix.tex
\appendix

\setcounter{tocdepth}{-1}
\addtocontents{toc}{\protect\setcounter{tocdepth}{2}}

\tableofcontents

\newpage

\section{Details of Methodology}

\subsection{Complete Algorithm Implementations}\label{sec:complet_algorithm}
We provide detailed implementations about training-free cluster-based routing (Algorithm~\ref{alg:cluster_routing_full}), and RL-driven Multi-step Routing (Algorithm~\ref{alg:rl_routing_full}).

\input{section/algorithm_training_free}
\input{section/algorithm_rl}

\subsection{Detailed Specification of Reward Signals}\label{sec:appendix_reward_signals}

To bridge the gap between structured interaction and task-specific accuracy, \textsc{Atlas} employs a composite reward function $r_\phi = \mathcal{R}_{\text{fmt}} + \gamma \mathcal{R}_{\text{out}} + \xi \mathcal{R}_{\text{sel}}$. This section provides the formal definitions and criteria for each reward component.

\paragraph{Format Reward ($\mathcal{R}_{\text{fmt}}$)}
The format reward ensures that the RL agent adheres to the predefined syntactic protocols, which is essential for stable parsing and environment interaction. $\mathcal{R}_{\text{fmt}}$ is set to $0$ if all the following conditions are satisfied, and $-1$ otherwise:
\begin{itemize}[left=0.5cm]
    \item \textbf{Tag Integrity:} All XML-style tags (e.g., \texttt{<think>}, \texttt{<route>}, and \texttt{<answer>}) must be correctly opened and closed in a nested or sequential manner.
    \item \textbf{Invocation Syntax:} Tool calls within the \texttt{search} block must strictly follow the format \texttt{Model-Name@@Tool-Name:Input}. Furthermore, the specified model and tool names must exist within the active routing pool $\mathcal{P}$.
    \item \textbf{Mandatory Reasoning:} The trajectory must contain at least one complete \texttt{<think>...</think>} block to ensure internal deliberation before an action or answer.
    \item \textbf{Uniqueness of Response:} The trajectory must conclude with exactly one \texttt{<answer>...</answer>} block.
    \item \textbf{Execution Consistency:} To maintain the integrity of the multi-step interaction, the number of \texttt{search} calls initiated by the agent must strictly match the number of \texttt{information} blocks returned by the environment.
\end{itemize}

\paragraph{Outcome Reward ($\mathcal{R}_{\text{out}}$)}
The outcome reward serves as the primary signal for task success. It is a binary indicator evaluated upon the completion of the trajectory:
\begin{equation}
\mathcal{R}_{\text{out}} = 
\begin{cases} 
1, & \text{if the answer } y_j \text{ is correct,} \\
0, & \text{otherwise.}
\end{cases}\label{eq_out}
\end{equation}

\paragraph{Model Selection Reward ($\mathcal{R}_{\text{sel}}$)}
To encourage the agent to select the most efficient and capable expert for a given domain, we introduce an alignment-based penalty. The ``optimal model'' for each task is pre-determined as follows:
\begin{itemize}
    \item For the \textbf{MBPP} dataset, the optimal model is defined as \texttt{Qwen2.5-Coder-7B-Instruct}.
    \item For the \textbf{Calculator} and \textbf{NQ} datasets, the optimal model is identified via an offline evaluation where \texttt{GPT-4o} judges the best-performing candidate from the pool for each specific query.
\end{itemize}
The reward is then formulated to penalize sub-optimal invocations:
\begin{equation}
\mathcal{R}_{\text{sel}} = 
\begin{cases} 
0, & \text{if select the optimal model,} \\
-0.15, & \text{otherwise.}
\end{cases}\label{eq_sel}
\end{equation}

\section{Additional Experimental Details}\label{sec:appendix_experimental_details}

\subsection{Datasets}\label{subsec:appendix_datasets}
The datasets utilized in this paper are summarized in Table \ref{table:dataset-details}. Below, we provide detailed descriptions of each benchmark to illustrate the diverse reasoning capabilities required by our framework.

\paragraph{Mathematical Reasoning.}
\begin{itemize}[leftmargin=1.36em]
    \item \textbf{AIME 2024 \& AIME 2025}~\citep{AIME2024, AIME2025}: The American Invitational Mathematics Examination (AIME) is a prestigious 15-question, 3-hour test designed for high-performing high school students. We evaluate on the 2024 and 2025 editions, each containing 30 problems that demand advanced problem-solving skills, strategic thinking, and precise numerical computation.
    
    \item \textbf{AMC}~\citep{lightman2023let}: The American Mathematics Competitions (AMC) consist of multiple-choice problems ranging from elementary to intermediate difficulty. Our evaluation set includes 40 problems that assess fundamental mathematical reasoning and computational proficiency.
\end{itemize}

\paragraph{Arithmetic Reasoning.}
\begin{itemize}[leftmargin=1.36em]
    \item \textbf{Calculator}~\citep{wu2025tool}: A benchmark containing 1,000 complex arithmetic problems requiring precise numerical computation. These problems test the model's ability to recognize when external calculation tools are necessary and to correctly formulate and interpret computational results, evaluating the integration of reasoning and tool invocation.
\end{itemize}

\paragraph{Code Generation.}
\begin{itemize}[leftmargin=1.36em]
    \item \textbf{HumanEval}~\citep{chen2021evaluating}: This benchmark comprises 164 hand-crafted programming problems designed to evaluate code synthesis capabilities. Each problem includes a function signature, docstring, body, and unit tests. Solutions require understanding natural language specifications and generating functionally correct Python code.
    
    \item \textbf{MBPP}~\citep{austin2021program}: The Mostly Basic Programming Problems (MBPP) dataset contains 974 crowd-sourced Python programming problems designed for entry-level programmers. Problems are described in natural language and require generating short Python functions, typically 1-10 lines of code. This benchmark tests basic programming constructs including loops, conditionals, and string manipulation.
\end{itemize}

\paragraph{Commonsense Reasoning.}
\begin{itemize}[leftmargin=1.36em]
    \item \textbf{Natural Questions (NQ)}~\citep{kwiatkowski2019natural}: A question-answering dataset containing real user queries issued to Google Search. Questions span diverse topics and require retrieving and synthesizing information from Wikipedia articles. This benchmark evaluates knowledge-intensive reasoning and information retrieval capabilities.
    
    \item \textbf{Web Questions (WebQ)}~\citep{webq}: A dataset of 1,000 questions designed to test knowledge-based question answering. Questions are sourced from web search queries and require retrieving factual information from knowledge bases, evaluating the model's ability to access and reason over external knowledge sources.
\end{itemize}

\paragraph{Logical Reasoning.}
\begin{itemize}[leftmargin=1.36em]
    \item \textbf{LogiQA2}~\citep{liu2023logiqa}: An improved version of LogiQA containing 1,572 multiple-choice logical reasoning problems. Questions are sourced from standardized exams and require identifying logical relationships, drawing inferences, and evaluating argument structures. This benchmark tests formal reasoning capabilities including deductive, inductive, and abductive reasoning.
\end{itemize}

\input{section/x_datasets}

\paragraph{Scientific Reasoning.}
\begin{itemize}[leftmargin=1.36em]
    \item \textbf{GPQA}~\citep{rein2024gpqa}: The Graduate-Level Google-Proof Q\&A benchmark consists of 448 multiple-choice questions across biology, physics, and chemistry, written by domain experts with PhD-level knowledge. Questions are designed to be difficult even for experts and require deep domain understanding beyond simple fact retrieval.
\end{itemize}

\paragraph{Multi-modal Reasoning.}
\begin{itemize}[leftmargin=1.36em]
    \item \textbf{ChartQA}~\citep{masry2022chartqa}: A visual question-answering benchmark containing questions about various chart types (bar charts, line graphs, pie charts). Questions require extracting quantitative information from visual representations and performing numerical reasoning, testing the integration of visual perception and mathematical computation.
    
    \item \textbf{Geometry3K}~\citep{lu2021inter}: A comprehensive geometry problem-solving dataset comprising multiple problems with diagram annotations. Problems involve diverse geometric concepts including angles, areas, perimeters, and spatial relationships. This benchmark evaluates visual-geometric reasoning and the ability to apply mathematical principles to diagrammatic representations.
    
    \item \textbf{TallyQA}~\citep{acharya2019tallyqa}: A visual counting dataset containing complex counting questions across diverse real-world images. Questions range from simple object counting to complex scenarios requiring spatial reasoning and selective attention. This benchmark tests fine-grained visual perception and numerical reasoning capabilities.
    
    \item \textbf{CountBench}~\citep{paiss2023teaching}: A specialized counting benchmark with questions designed to evaluate precise object enumeration in images. Unlike traditional counting tasks, CountBench emphasizes accuracy on challenging cases involving occlusions, similar objects, and cluttered scenes, requiring robust visual understanding.

    \item \textbf{TableVQA}~\citep{kim2024tablevqa}: A visual question-answering benchmark containing questions about tables across multiple domains. Questions require understanding table structures, extracting relevant information, and performing reasoning over tabular data, evaluating the integration of visual perception and structured data comprehension.

\end{itemize}

These diverse benchmarks collectively assess the framework's ability to dynamically select optimal model-tool combinations across varying task requirements, ranging from symbolic mathematical reasoning to multi-modal visual understanding.

\subsection{Baselines}
In our experiments, we compare the proposed methods against six baseline approaches. Below, we provide detailed descriptions of each baselines.
\begin{itemize}
    \item \textbf{Zero-shot (ZS) Router}: A baseline that directly prompts a base LLM to select the most suitable candidate model-tool combination from the available pool without prior examples.
    \item \textbf{Few-shot (FS) Router}: An extension of the zero-shot approach that incorporates several in-context examples to provide the base LLM with task-specific demonstrations and routing guidance.
    \item \textbf{Random Router}: A stochastic baseline that selects a candidate model-tool combination uniformly at random from the candidate pool for each query.
    \item \textbf{RouterDC} \cite{chen2024routerdc}: A routing framework based on dual contrastive learning that maps queries and model-tool combinations into a shared embedding space. It utilizes sample-LLM and sample-sample contrastive losses to optimize query-model alignment and selects the optimal combination via cosine similarity.
    \item \textbf{MLPRouter} \cite{hu2024routerbench}: A classification-based framework that trains an MLP for each model-tool combination. Each MLP predicts the success probability of its corresponding combination, and the one with the highest output is selected.
    \item \textbf{BertRouter} \cite{ong2024routellm}: A router utilizing a pre-trained mDeBERTaV3-base encoder \cite{he2021debertav3} with an integrated classification head to predict the accuracy of model-tool pairings, following a selection logic similar to MLPRouter.
\end{itemize}

\subsection{Evaluation Details.}\label{subsec:appendix_evaluation_details}
Our experiments employ two evaluation protocols: \textbf{In-Distribution (ID)}, where each dataset has its own training split, and \textbf{Out-of-Distribution (OOD)}, where models are trained exclusively on three datasets (Calculator, NQ, MBPP) and evaluated on all ten benchmarks, making AIME24, AIME25, AMC, HumanEval, WebQ, LogiQA2, and GPQA fully out-of-domain. For cluster-based routing in OOD settings, semantic clusters and performance statistics are derived \textit{solely} from the three training datasets; test queries from unseen domains are assigned to the nearest cluster based on semantic similarity, without accessing any OOD test set information. This design reflects realistic domain-specific scenarios but inevitably suffers from cluster misalignment on unfamiliar tasks (49.2\% OOD vs. 63.5\% ID, Table~\ref{tab:main_results}). In contrast, RL-based routing learns transferable patterns (when to invoke symbolic tools or defer to specialized models) that generalize beyond training distributions, achieving 59.4\% OOD accuracy. Importantly, no test set information is leaked: all routing decisions rely purely on query embeddings and training domain statistics, ensuring evaluation integrity and demonstrating that gains stem from our dual-path architecture's complementary strengths.

\input{section/appendix_main_results}

\subsection{Implementation Details}\label{subsec:appendix_implementation_details}
\paragraph{Hyperparameters for Cluster-based Routing.}
We set the number of cluster centers to 8 and employ the KMeans algorithm with the following hyperparameters: the cluster centers are initialized using the k-means++ method to accelerate convergence; the algorithm is allowed up to 1000 iterations per run; the number of initializations is set to automatic selection, the hyperparameter $\alpha$ in Equation~\ref{eq_acc_cost} is set to 0.5; and the Elkan variant of KMeans is used for computational efficiency.

\begin{figure*}[t]
  \includegraphics[width=0.98\linewidth]{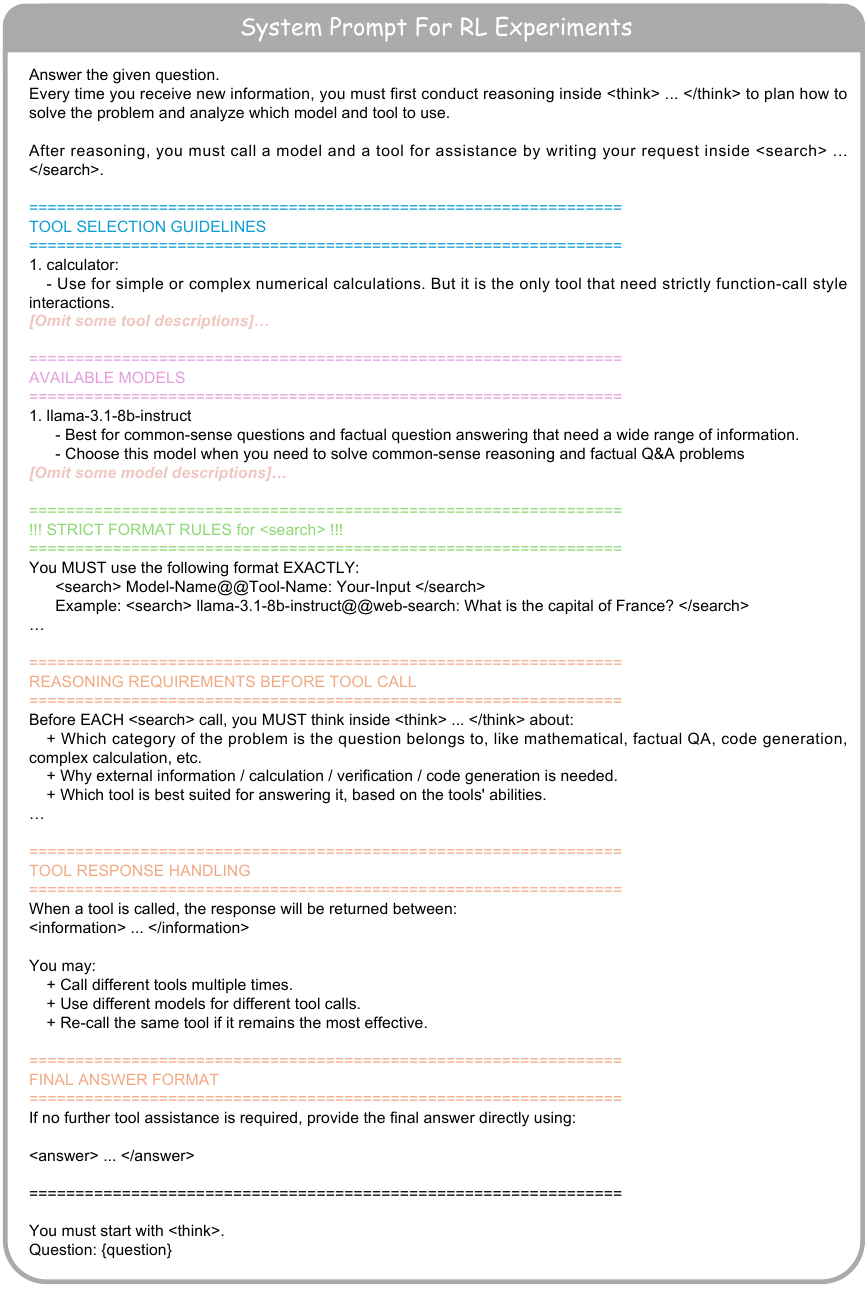}
  \caption{System prompt for \textsc{Atlas} RL Experiments.}
  \label{fig:system_prompt}
\end{figure*}

\paragraph{Hyperparameters for RL Training.}
We train the policy model (Qwen2.5-3B-Instruct) using PPO with generalized advantage estimation (GAE). The training and validation batch sizes are both set to 24. The maximum prompt length is 4096 tokens, while the maximum response length is set to 3000 tokens. To control context growth, the maximum lengths for the observations are set to 2048 tokens each, and the maximum number of interaction turns is limited to 4. The actor is optimized with a learning rate of $1\times10^{-6}$, while the critic uses a learning rate of $1\times10^{-5}$. The PPO mini-batch size and micro-batch size for the actor are set to 12 and 6, respectively. The KL-divergence coefficient is fixed to 0.001. During rollout, we use a temperature of 1.0. For the reward weights in $r_\phi = \mathcal{R}_{\text{fmt}} + \gamma \mathcal{R}_{\text{out}} + \xi \mathcal{R}_{\text{sel}}$, we assign $\gamma=\xi=1$. All experiments are conducted for 250 total training steps. We also provide the RL system prompt in Figure~\ref{fig:system_prompt}.

\begin{figure*}[t]
  \includegraphics[width=0.98\linewidth]{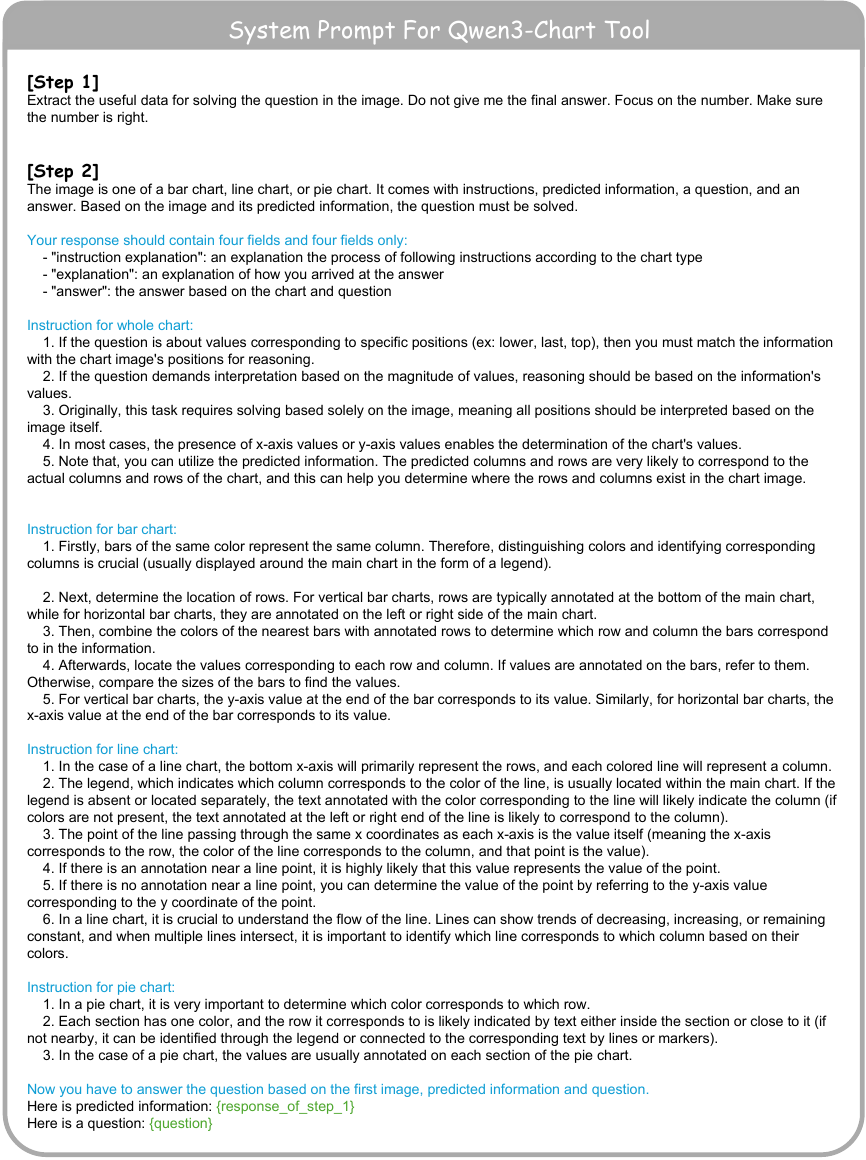}
  \caption{System prompt for Qwen3-Chart Tool.}
  \label{fig:tool_chart}
\end{figure*}

\begin{figure*}[t]
  \includegraphics[width=0.98\linewidth]{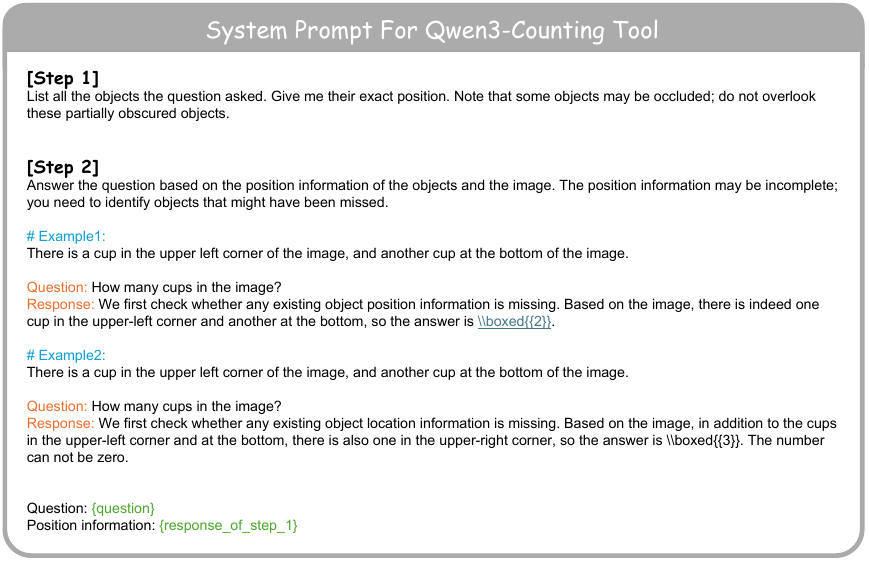}
  \caption{System prompt for Qwen3-Counting Tool.}
  \label{fig:tool_counting}
\end{figure*}

\begin{figure*}[t]
  \includegraphics[width=0.98\linewidth]{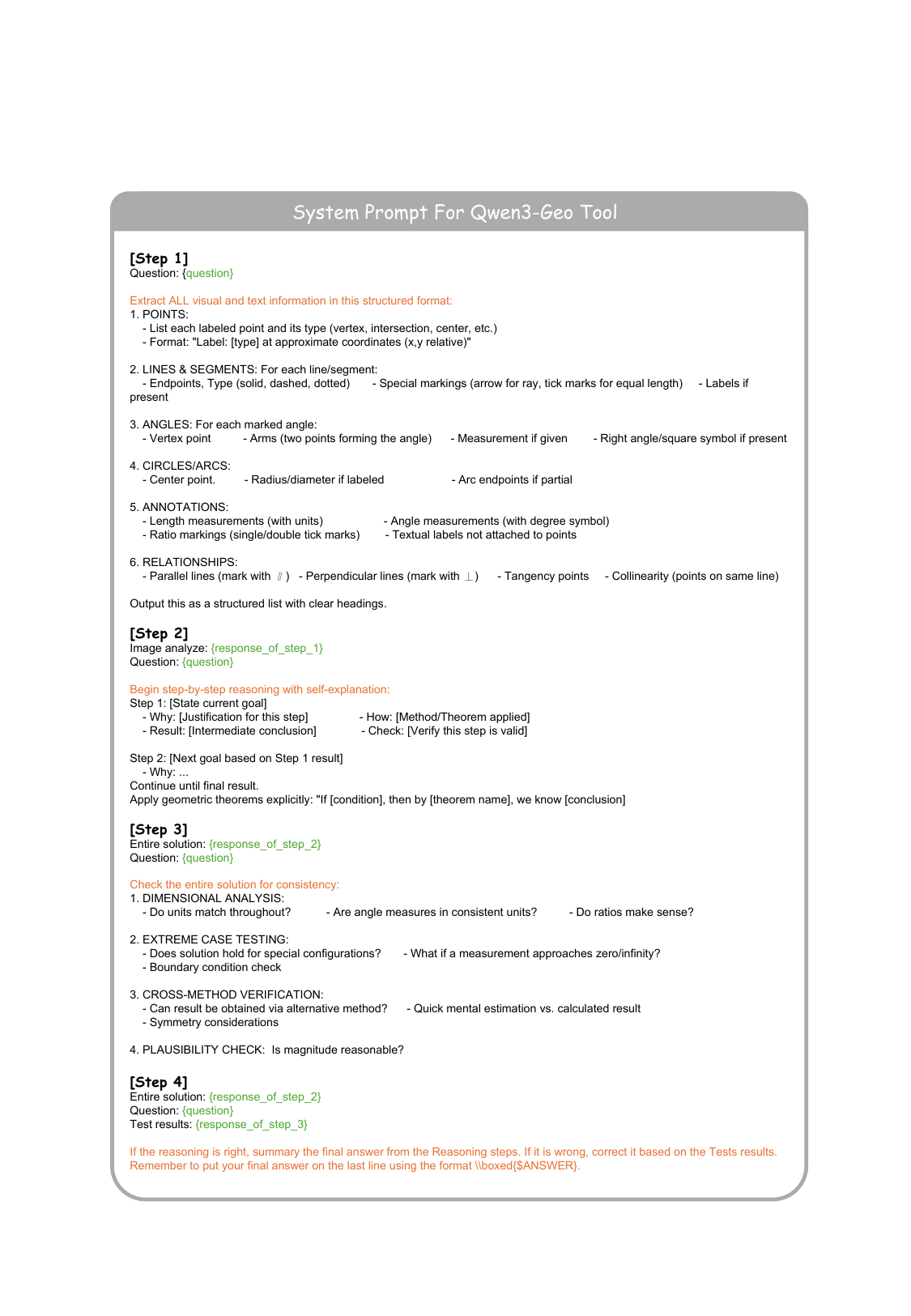}
  \caption{System prompt for Qwen3-Counting Tool.}
  \label{fig:tool_geo}
\end{figure*}

\paragraph{Tool Details.}
We provide the system prompt for three special multimodal tools in Figure~\ref{fig:tool_chart}-\ref{fig:tool_geo}. Regarding text-based tools, the \textbf{\textit{Code Interpreter}} executes Python code, returns the execution results, indicates whether the execution was successful, and reports error locations and underlying causes in case of failures. The \textbf{\textit{Web Search}} tool leverages the official Google Custom Search API to retrieve the three most relevant search result snippets. Search results are obtained by sending HTTP GET requests to the API (\url{https://www.googleapis.com/customsearch/v1}) with the required parameters, including the API key, search engine ID, query string, and the number of top results to return. The \textbf{\textit{Calculator}} parses the model output in a function-call format to extract the mathematical expression, computation type, and precision requirements, and then computes and returns the result using appropriate functions from the \textit{sympy} library. The \textbf{\textit{Process Reward Model (PRM)}} runs five model outputs in parallel, evaluates the segmented outputs using a reward model, and selects the output with the highest average score as the final result. In this work, we use the off-the-shelf Qwen2.5-Math-PRM-7B\footnote{\href{https://huggingface.co/Qwen/Qwen2.5-Math-PRM-7B}{Qwen/Qwen2.5-Math-PRM-7B}}.

\paragraph{Baseline Details.}
The original baselines perform routing among multiple models. When evaluating these baselines, we replace the models in the code with model–tool combinations, thereby enabling the baselines to route over both models and tools simultaneously. For example, when reproducing EmbedLLM~\citep{zhuang2025embedllm}, we substitute the model names in the training data with the model–tool combination names, and replace the model performance with the empirically measured performance of the model–tool combinations. Apart from these modifications, all training and evaluation procedures strictly follow the official open-source implementations, ensuring that the reported results faithfully reflect the true performance of the baseline methods. When evaluating closed-source models, we use exactly the same evaluation code and prompt templates (including the use of CoT reasoning and fixed answer formats) as those used for other baselines and our proposed method, ensuring a strictly fair comparison and convincing final results.

\begin{table*}[t]
    \small
    \centering
    \caption{\textbf{Performance comparison of \textsc{Atlas} against single-tool baselines across multi-modal benchmarks}. The framework dynamically routes queries among multi-modal tools using Qwen3-VL-8B-Instruct as the backbone. `None' represents direct reasoning without any tools. The best results are highlighted in \textbf{bold}. }
    \label{tab:multimodal}
    \begin{tabular}{lcccccc}
    \toprule
    \multirow{2}{*}{\textbf{Tool}} & \multicolumn{2}{c}{\textbf{Chart Understanding}} & \multicolumn{1}{c}{\textbf{Math Reasoning}} & \multicolumn{2}{c}{\textbf{Object Enumeration}} & \multirow{2}{*}{\textbf{Avg.}} \\
    \cmidrule(lr){2-3} \cmidrule(lr){4-4} \cmidrule(lr){5-6}
    & ChartQA & TableVQA & Geometry3K & TallyQA & CountBench & \\
    \midrule
    None (Direct Reasoning) & 81.2 & 64.5 & 57.9 & 23.9 & 84.1 & 62.3 \\
    \;\;+OCR               & 79.6 & 67.0 & 57.4 & 27.9 & 83.9 & 63.2 \\
    \;\;+Qwen3-Chart       & 83.0 & 62.4 & 50.2 & 32.5 & 87.8 & 63.2 \\
    \;\;+Qwen3-Counting    & 80.0 & 64.8 & 54.4 & 34.1 & 89.6 & 64.6 \\
    \;\;+Qwen3-Geo         & 75.8 & 62.4 & 58.7 & 32.9 & 87.8 & 63.5 \\
    \midrule
    \rowcolor[RGB]{236,244,252}
    \textbf{\textsc{Atlas} (ours)}    & \textbf{84.0} & \textbf{68.2} & \textbf{65.6} & \textbf{36.7} & \textbf{90.2} & \textbf{68.9} \\
    \rowcolor[RGB]{245, 238, 248}
    $\bigtriangleup$ vs. None    & +2.8 & +3.7 & +7.7 & +12.8 & +6.1 & +6.6 \\
    \bottomrule
    \end{tabular}
\end{table*}

\paragraph{Computing Details.}
All experiments are conducted on eight NVIDIA A100-80GB GPUs. 

\section{More Results and Analysis}\label{sec:appendix_additional_results}
\subsection{Details Main Results}\label{appendix_main_results}
We provide extended comparisons in Table~\ref{tab:appendix_main_results}, incorporating additional closed-source models (e.g., Gemini-2.5-Flash) and routing baselines including GraphRouter~\citep{feng2025graphrouter} and EmbedLLM~\citep{zhuang2025embedllm}.

\subsection{Detailed Multimodal Results}\label{appendix_multimodal}
Visual perception, comprehension, and reasoning are crucial capabilities for autonomous agents~\citep{xie2024large,wu2025boosting,reddy2025orion}. We conduct multimodal extension experiments as described in Section~\ref{subsec:multi_modal}, with detailed orchestration results provided in Table~\ref{tab:multimodal}. Our evaluation spans diverse visual reasoning tasks: chart understanding (ChartQA~\citep{masry2022chartqa}, TableVQA~\citep{kim2024tablevqa}), math reasoning (Geometry3K~\citep{lu2021inter}), and object enumeration (TallyQA~\citep{acharya2019tallyqa}, CountBench~\citep{paiss2023teaching}).

To ensure fair comparison, all configurations, including the baseline (direct inference with Qwen3-8B-VL), single-tool baselines, and \textsc{Atlas}, share the same foundational model (Qwen3-8B-VL) and identical evaluation protocols. The only modification across settings is the inclusion of tool-invocation instructions in the system prompt, which guide the model on when and how to invoke specific tools (e.g., chart parsing, object counting, geometric reasoning). Crucially, the core reasoning capacity and model parameters remain unchanged, ensuring that observed performance gains stem from adaptive tool orchestration rather than model-level differences or prompt engineering artifacts. The results demonstrate that \textsc{Atlas} consistently outperforms single-tool baselines across all categories, validating the effectiveness of dynamic tool orchestration in multimodal scenarios. We plan to explore more backbones in future work.

\begin{table}[t]
\small
\centering
\caption{Performance scaling with Self-Consistency (SC) across different sample sizes.}
\label{tab:sc_scaling}
\begin{tabular}{lccccc}
\toprule
\textbf{Dataset} & \textbf{Pass@1} & \textbf{SC@4} & \textbf{SC@8} & \textbf{SC@16} \\
\midrule
AIME24 & 43.3 & 63.3 & 66.7 & \cellcolor[RGB]{245, 238, 248}{70.0} \\
AIME25 & 40.0 & 43.3 & 46.7 & \cellcolor[RGB]{245, 238, 248}{50.0} \\
AMC & 82.5 & 92.5 & 95.0 & \cellcolor[RGB]{245, 238, 248}{97.5} \\
Calc. & 83.3 & 83.5 & 84.7 & \cellcolor[RGB]{245, 238, 248}{86.9} \\
GPQA & 46.4 & 53.6 & 57.1 & \cellcolor[RGB]{245, 238, 248}{59.4} \\
\bottomrule
\end{tabular}
\end{table}

\begin{table*}[t]
    \small
    \centering
    \caption{\textbf{Distribution of dominant model-tool combinations across diverse benchmarks}. Dominant combination indicates the most frequently selected model-tool pair by our framework for each specific dataset.}
    \label{tab:dominant}
    \begin{tabular}{lccccc}
    \toprule
    \textbf{Dataset} & \textbf{Dominant Combination} & \textbf{\textsc{Atlas}(Cluster)} & \textbf{Dominant Combination} & \textbf{\textsc{Atlas}(RL)} \\
    \midrule
    AIME24 & \cellcolor[RGB]{236,244,252}{DeepSeek.-7B@PRM}      & \cellcolor[RGB]{245, 238, 248}{100.0\%}  & \cellcolor[RGB]{236,244,252}{DeepSeek.-7B@PRM}    & \cellcolor[RGB]{245, 238, 248}{100.0\%} \\
    AIME25 & \cellcolor[RGB]{236,244,252}{DeepSeek.-7B@PRM}      & \cellcolor[RGB]{245, 238, 248}{100.0\%}  & \cellcolor[RGB]{236,244,252}{DeepSeek.-7B@PRM}    & \cellcolor[RGB]{245, 238, 248}{100.0\%} \\
    AMC    & \cellcolor[RGB]{236,244,252}{DeepSeek.-7B@PRM}      & \cellcolor[RGB]{245, 238, 248}{100.0\%}  & \cellcolor[RGB]{236,244,252}{DeepSeek.-7B@PRM}    & \cellcolor[RGB]{245, 238, 248}{91.7\%}  \\
    Human. & \cellcolor[RGB]{236,244,252}{Coder-7B@Python}       & \cellcolor[RGB]{245, 238, 248}{100.0\%}  & \cellcolor[RGB]{236,244,252}{Coder-7B@Python}     & \cellcolor[RGB]{245, 238, 248}{100.0\%} \\
    MBPP   & \cellcolor[RGB]{236,244,252}{Coder-7B@Python}       & \cellcolor[RGB]{245, 238, 248}{100.0\%}  & \cellcolor[RGB]{236,244,252}{Coder-7B@Python}     & \cellcolor[RGB]{245, 238, 248}{100.0\%} \\
    Calc.  & \cellcolor[RGB]{236,244,252}{Qwen2.5-7B@Calc.}      & \cellcolor[RGB]{245, 238, 248}{100.0\%}  & \cellcolor[RGB]{236,244,252}{Qwen2.5-7B@Calc.}    & \cellcolor[RGB]{245, 238, 248}{95.8\%}  \\
    NQ     & \cellcolor[RGB]{236,244,252}{Llama3.1-8B@Search}    & \cellcolor[RGB]{245, 238, 248}{92.8\%}   & \cellcolor[RGB]{236,244,252}{Llama3.1-8B@Search}  & \cellcolor[RGB]{245, 238, 248}{99.0\%}  \\
    WebQ   & \cellcolor[RGB]{236,244,252}{Llama3.1-8B@Search}    & \cellcolor[RGB]{245, 238, 248}{98.8\%}   & \cellcolor[RGB]{236,244,252}{Llama3.1-8B@Search}  & \cellcolor[RGB]{245, 238, 248}{100.0\%} \\
    LQA2   & \cellcolor[RGB]{236,244,252}{InternLM3-8B@Search}   & \cellcolor[RGB]{245, 238, 248}{99.7\%}   & \cellcolor[RGB]{236,244,252}{InternLM3-8B@Search} & \cellcolor[RGB]{245, 238, 248}{56.4\%}  \\
    GPQA   & \cellcolor[RGB]{236,244,252}{DeepSeek.-7B@Python}   & \cellcolor[RGB]{245, 238, 248}{80.4\%}   & \cellcolor[RGB]{236,244,252}{DeepSeek.-7B@Python} & \cellcolor[RGB]{245, 238, 248}{95.5\%}  \\
    \bottomrule
    \end{tabular}
\end{table*}

\subsection{Test-Time Scaling Results}\label{appendix_tts}
We analyze the scalability of our approach by increasing the self-consistency (SC) sample count on several representative benchmarks. As illustrated in Table \ref{tab:sc_scaling}, performance across almost all datasets shows a positive correlation with the number of samples ($k$). For example, on the AIME24 benchmark, SC@16 yields a significant improvement from 43.3\% to 70.0\%. Similar findings are also observed on other tasks, such as commonsense reasoning and scientific reasoning. These results demonstrate that the ensemble of model-tool combinations provides a more robust candidate pool for majority voting.

\begin{table*}[h]
\centering
\caption{\textbf{Performance and latency under different tool pool sizes.}}
\label{tab:toolpool_scale}
\small
\begin{tabular}{cccccccc}
\toprule
Number of Tools & AIME25 & AMC & MBPP & Calc. & LQA2 & GPQA & Avg. \\
\midrule
\multicolumn{8}{l}{\textit{Accuracy (\%)}} \\
2 & 23.3 & 60.0 & 72.3 & 56.1 & 60.3 & 40.2 & \cellcolor[RGB]{236,244,252}52.0 \\
4 & 33.3 & 67.5 & 81.8 & 81.6 & 62.7 & 42.0 & \cellcolor[RGB]{236,244,252}61.5 \\
\rowcolor[RGB]{245, 238, 248}
8 & \textbf{36.7} & \textbf{70.0} & \textbf{81.8} & \textbf{83.3} & \textbf{62.9} & \textbf{42.9} & \cellcolor[RGB]{236,244,252}\textbf{62.9} \\
\midrule
\multicolumn{8}{l}{\textit{Time Latency per Query (s)}} \\
2 & 67.96 & 34.16 & 3.91 & 0.34 &  1.28 & 75.96 & \cellcolor[RGB]{236,244,252}30.6 \\
4 & 123.15& 45.20 & 4.02 & 0.26 &  1.35 & 84.68 & \cellcolor[RGB]{236,244,252}43.1 \\
\rowcolor[RGB]{245, 238, 248}
8 & 132.62& 44.84 & 4.85 & 0.42 &  1.65 & 86.95 & \cellcolor[RGB]{236,244,252}45.2 \\
\bottomrule
\end{tabular}
\end{table*}

\subsection{Analysis of Model-Tool Alignment Preferences}\label{appendix_invocation_preference}
Table \ref{tab:dominant} illustrates the strategic alignment between specific models and tools across diverse benchmarks. In deterministic domains such as coding (HumanEval) and advanced mathematics (AIME), \textsc{Atlas} exhibits a clear convergence, selecting specialized pairings like Qwen2.5-Coder-7B with Python or DeepSeek-R1 with PRM in nearly 100\% of cases. This high degree of consistency confirms the framework's ability to internalize the performance advantages of domain-specific modules.

In contrast, knowledge-intensive tasks (NQ, MedQA) trigger a transition toward retrieval-augmented configurations, primarily utilizing Llama-3.1-8B with Web-Search. For more complex, broad-spectrum benchmarks like LQA2, the selection distribution becomes significantly more granular, with the dominant combination of \textsc{Atlas}(RL) accounting for only 56.4\%. This shift demonstrates that \textsc{Atlas} avoids rigid heuristics, instead employing a flexible orchestration strategy that adapts to the specific nuances and difficulty of each query.

\subsection{Scalability of RL-Based Routing with Tool Pool Size}\label{app:toolpool_scale}
As shown in Table~\ref{tab:toolpool_scale}, we evaluate how the RL routing policy scales as the tool pool grows from 2 to 8 tools:
\begin{itemize}[leftmargin=*]
  \item \textbf{2 Tools}: Web-search, Python-code.
  \item \textbf{4 Tools}: Above + Calculator, PRM.
  \item \textbf{8 Tools}: Above + Reasoning Chain Refiner,
        Tool-Use Backtracer, Code Debugger, Segment Refiner.
\end{itemize}

As tool pool size grows from 4 to 8, accuracy improves by $+1.5\%$ on average while latency increases by only $\sim$2\,s, indicating \textbf{sub-linear latency growth}. This demonstrates that the RL policy effectively identifies relevant tools even as the action space expands, owing to the transferable routing principles learned during training.

\subsection{Evaluation on Realistic Tool-Calling Benchmark (BFCL)}
\label{app:bfcl}
We evaluate \textsc{Atlas}(RL) on the Berkeley Function Calling Leaderboard (BFCL)~\cite{patil2025the}, a benchmark specifically designed for realistic, production-oriented tool-invocation scenarios. As shown in Table~\ref{tab:bfcl}, we report single-round results across four sub-categories.

\begin{table*}[h]
\centering
\caption{\textbf{BFCL single-round tool invocation results}.}
\label{tab:bfcl}
\small
\begin{tabular}{lccccr}
\toprule
Method & Simple & Multiple & Parallel & Multiple Parallel & Avg. \\
\midrule
o3                    & 74.3 & 89.0 & 86.5 & 78.0 & 81.9 \\
GPT-5.2               & 72.9 & 88.0 & 89.0 & 77.5 & 81.9 \\
Grok-4                & 67.0 & 93.5 & 89.0 & 87.5 & 84.3 \\
Gemini-2.5-Flash      & 74.3 & 92.0 & 94.0 & 79.5 & 85.0 \\
\rowcolor[RGB]{245, 238, 248}
\textbf{\textsc{Atlas}(RL)} 
                      & \textbf{76.0} & \textbf{93.5} & \textbf{91.0} 
                      & \textbf{83.5} & \textbf{86.0} \\
\bottomrule
\end{tabular}
\end{table*}

\textsc{Atlas}(RL) achieves an average score of 86.0\%, surpassing all listed closed-source models and validating the effectiveness of our approach in complex, real-world tool-calling scenarios.

\begin{table*}[t]
\small
\centering
\caption{\textbf{Ablation study on reward components.} We evaluate the impact of removing $\mathcal{R}_{\text{sel}}$ (model selection reward) and $\mathcal{R}_{\text{fmt}}$ (format reward) on out-of-distribution performance. Models are trained on Calc., NQ, and MBPP ($\ddag$), then evaluated on all datasets. The best results are highlighted in \textbf{bold}.}
\label{tab:appendix_ablation_model_selection}
\resizebox{1.0\linewidth}{!}{
\begin{tabular}{lccccccccccc}
\toprule
\multirow{2}{*}{\textbf{Method}} & \multicolumn{3}{c}{\textbf{Math Reasoning}} & \multicolumn{2}{c}{\textbf{Code}} & \textbf{Arith.} & \multicolumn{2}{c}{\textbf{Common.}} & \textbf{Logic} & \textbf{Sci.} & \multirow{2}{*}{\textbf{Avg.}} \\
\cmidrule(lr){2-4} \cmidrule(lr){5-6} \cmidrule(lr){7-7} \cmidrule(lr){8-9} \cmidrule(lr){10-10} \cmidrule(lr){11-11}
& AIME24 & AIME25 & AMC & Human. & MBPP$^\ddag$ & Calc.$^\ddag$ & NQ$^\ddag$ & WebQ & LQA2 & GPQA & \\
\midrule
\rowcolor{gray!8}\multicolumn{12}{c}{\textit{Training-free Baselines}} \\
ZS Router        & 13.3 & 6.7  & 32.5 & 53.0 & 64.2 & 55.7 & 29.2 & 39.2 & 45.3 & 24.6 & 36.4 \\
FS Router        & 23.3 & 13.3 & 40.0 & 68.9 & 64.7 & 47.2 & 27.3 & 35.8 & 40.8 & 25.9 & 38.7 \\
Random Router    & 6.7  & 3.3  & 15.0 & 37.8 & 52.6 & 40.2 & 25.3 & 32.1 & 49.2 & 30.6 & 29.3 \\
\midrule
\rowcolor{gray!8}\multicolumn{12}{c}{\textit{Training-based Baselines}} \\
RouterDC         & 13.3 & 3.3  & 47.5 & 79.2 & 78.7 & 70.8 & 40.1 & 50.8 & 50.4 & 28.6 & 46.3 \\
GraphRouter      & 16.7 & 3.3 & 42.5 & 76.2 & 73.4 & 71.2 & 36.5 & 49.3 & 47.2 & 27.7 & 44.4 \\
EmbedLLM         & 13.3 & 3.3 & 45.0 & 79.9 & 73.0 & 79.1 & 41.4 & 50.2 & 51.5 & 31.7 & 46.8 \\
MLPRouter        & 13.3 & 3.3  & 32.5 & 75.0 & 67.7 & 54.6 & 37.3 & 43.7 & 38.9 & 26.8 & 39.3 \\
BertRouter       & 6.7  & 6.7  & 40.0 & 78.7 & 79.0 & 67.0 & 38.9 & 51.4 & 40.3 & 27.7 & 43.6 \\
\midrule
\midrule
\rowcolor[RGB]{236,244,252}
\textbf{\textsc{Atlas} (RL)} & \textbf{43.3} & \textbf{33.3} & \textbf{67.5} & \textbf{85.4} & \textbf{81.8} & \textbf{81.6} & \textbf{44.1} & \textbf{52.2} & 62.7 & \textbf{42.0} & \textbf{59.4} \\
\;\;\;w/o $\mathcal{R}_{\text{sel}}$ & 36.7 & 26.7 & 65.0 & 82.3 & 80.6 & 79.1 & 41.3 & 48.3 & \textbf{62.9} & 40.6 & 56.3 \\ 
\rowcolor[RGB]{245,238,248}
\;\;\;$\bigtriangleup$ & -6.6 & -6.6 & -2.5 & -3.1 & -1.2 & -2.5 & -2.8 & -3.9 & +0.2 & -1.4 & -3.1 \\ 
\midrule
\rowcolor[RGB]{236,244,252}
\textbf{\textsc{Atlas} (RL)} & \textbf{43.3} & \textbf{33.3} & \textbf{67.5} & \textbf{85.4} & \textbf{81.8} & \textbf{81.6} & \textbf{44.1} & \textbf{52.2} & \textbf{62.7} & \textbf{42.0} & \textbf{59.4} \\
\;\;\;w/o $\mathcal{R}_{\text{fmt}}$ & 33.3 & 26.7 & 55.0 & 78.0 & 75.4 & 78.3 & 41.6 & 48.0 & 58.2 & 38.4 & 53.3 \\ 
\rowcolor[RGB]{245,238,248}
\;\;\;$\bigtriangleup$ & -10.0 & -6.6 & -12.5 & -7.4 & -6.4 & -3.3 & -2.5 & -4.2 & -4.5 & -3.6 & -6.1 \\ 
\bottomrule
\end{tabular}}
\end{table*}

\subsection{Ablation Study on Reward Components}\label{appendix_ablation_study}
To investigate individual reward contributions, we train the RL policy without $\mathcal{R}_{\text{sel}}$ or $\mathcal{R}_{\text{fmt}}$ while keeping other signals intact. This addresses concerns about potential circularity from GPT-4o judgments in $\mathcal{R}_{\text{sel}}$ and validates the necessity of format enforcement.

As shown in Table~\ref{tab:appendix_ablation_model_selection}, removing $\mathcal{R}_{\text{sel}}$ causes a modest 3.1\% degradation (59.4\% $\rightarrow$ 56.3\%), with notable drops on mathematical reasoning (AIME24/25: $-$6.6\% each). However, the policy still substantially outperforms all baselines, including RouterDC (46.3\%) and EmbedLLM (46.8\%). The retained performance (56.3\% vs. 36.4\% for zero-shot) confirms that \textsc{Atlas} learns effective routing independently through $\mathcal{R}_{\text{fmt}}$ and $\mathcal{R}_{\text{out}}$, without requiring external model judgments. This validates $\mathcal{R}_{\text{sel}}$ as an efficiency-oriented auxiliary signal rather than a necessary component.

In contrast, removing $\mathcal{R}_{\text{fmt}}$ leads to a more substantial 6.1\% degradation (59.4\% $\rightarrow$ 53.3\%), with significant drops on mathematical reasoning (AIME24: $-$10.0\%, AMC: $-$12.5\%) and code generation (HumanEval: $-$7.4\%). This reveals that format enforcement is critical for maintaining structured interaction patterns—proper tool syntax and reasoning-action sequencing—which form the foundation for multi-step orchestration. Without $\mathcal{R}_{\text{fmt}}$, the policy produces malformed tool calls that propagate failures throughout reasoning trajectories. These results validate our design: $\mathcal{R}_{\text{fmt}}$ and $\mathcal{R}_{\text{out}}$ constitute essential signals, while $\mathcal{R}_{\text{sel}}$ provides optional efficiency guidance.

\begin{table*}[t]
\small
\centering
\caption{\textbf{Sensitivity analysis on cluster number $K$.} Performance across different cluster granularities in cluster-based routing. All datasets have training data available (in-distribution setting).}
\label{tab:Sensitivity_analysis_on_cluster_number}
\resizebox{1.0\linewidth}{!}{
\begin{tabular}{lccccccccccc}
\toprule
\multirow{2}{*}{\textbf{$\#$Cluster}} & \multicolumn{3}{c}{\textbf{Math Reasoning}} & \multicolumn{2}{c}{\textbf{Code}} & \textbf{Arith.} & \multicolumn{2}{c}{\textbf{Common.}} & \textbf{Logic} & \textbf{Sci.} & \multirow{2}{*}{\textbf{Avg.}} \\
\cmidrule(lr){2-4} \cmidrule(lr){5-6} \cmidrule(lr){7-7} \cmidrule(lr){8-9} \cmidrule(lr){10-10} \cmidrule(lr){11-11}
& AIME24 & AIME25 & AMC & Human. & MBPP$^\ddag$ & Calc.$^\ddag$ & NQ$^\ddag$ & WebQ & LQA2 & GPQA & \\
\midrule
4 & 36.7 & 30.0 & 75.0 & 43.3 & 71.5 & 79.1 & 28.8 & 48.5 & 66.8 & 39.6 & 51.9 \\
\rowcolor[RGB]{245,238,248}
8 & \textbf{43.3} & \textbf{40.0} & \textbf{82.5} & \textbf{91.5} & \textbf{83.6} & \textbf{83.3} & \textbf{43.8} & \textbf{53.6} & \textbf{66.8} & \textbf{46.4} & \textbf{63.5} \\
16 & 40.0 & \textbf{40.0} & \textbf{82.5} & 90.9 & 82.9 & 82.3 & 44.1 & 53.4 & 66.7 & 45.3 & 62.8 \\
\bottomrule
\end{tabular}}
\end{table*}

\begin{table*}[h]
\centering
\caption{In-distribution performance across embedding models (cluster-based routing).}
\label{tab:embed_sensitivity}
\small
\begin{tabular}{lcccccr}
\toprule
Embedding Model & AIME25 & AMC & MBPP & LQA2 & GPQA & Avg. \\
\midrule
\texttt{embedding-gemma-300m}       & 40.0 & 82.5 & 83.8 & 66.2 & 46.2 & \cellcolor[RGB]{236,244,252}63.7 \\
\texttt{jina-embeddings-v3}         & 40.0 & 80.0 & 83.7 & 66.2 & 45.5 & \cellcolor[RGB]{236,244,252}63.1 \\
\texttt{nomic-embed-text-v1.5}      & 40.0 & 82.5 & 83.7 & 66.1 & 43.1 & \cellcolor[RGB]{236,244,252}63.1 \\
\texttt{gte-Qwen2-1.5B-instruct}    & 40.0 & 82.5 & 83.8 & 66.3 & 46.9 & \cellcolor[RGB]{236,244,252}63.9 \\
\rowcolor[RGB]{245, 238, 248}
\texttt{gte-Qwen2-7B-instruct} (ours) & 40.0 & 82.5 & 83.6 & 66.8 & 46.4 & \cellcolor[RGB]{236,244,252}63.9 \\
\bottomrule
\end{tabular}
\end{table*}

\subsection{Sensitivity Analysis on Cluster Number}\label{appendix_sensitivity_analysis}
To evaluate the robustness of cluster-based routing to the choice of cluster granularity, we conduct sensitivity analysis by varying the number of clusters $K \in \{4, 8, 16\}$ while keeping all other hyperparameters fixed. As shown in Table~\ref{tab:Sensitivity_analysis_on_cluster_number}, the optimal performance is achieved at $K=8$ with 63.5\% average accuracy, representing an 11.6\% improvement over $K=4$ (51.9\%) and a modest 0.7\% gain over $K=16$ (62.8\%). The substantial performance drop at $K=4$ suggests that overly coarse clustering fails to capture fine-grained task distinctions, leading to suboptimal model-tool alignments, particularly evident on code generation (HumanEval: 43.3\% vs. 91.5\%) where diverse programming patterns require more specialized routing. Conversely, increasing to $K=16$ yields diminishing returns, as excessively fine-grained clusters may suffer from data sparsity within each partition, resulting in less reliable performance statistics. These results demonstrate that moderate cluster granularity ($K=8$) strikes an effective balance between semantic specificity and statistical robustness, though the framework remains reasonably stable across a range of $K$ values (62.8\%–63.5\% for $K \in \{8, 16\}$), indicating limited sensitivity to this hyperparameter in practical deployments.

\subsection{Sensitivity of Cluster-Based Routing to Embedding Model}
\label{app:embed_sensitivity}
As shown in Table~\ref{tab:embed_sensitivity}, we evaluate five embedding models spanning 300M--7B parameters to assess whether the cluster-based routing is sensitive to encoder choice.

Performance variance across models is minimal (63.1\%--63.9\%), demonstrating that the cluster-based routing is \textbf{fundamentally robust to embedding model choice}. This stability arises because semantic clustering depends primarily on the quality of the semantic embedding space rather than encoder-specific characteristics: as long as the encoder captures task-level semantics, the downstream cluster-to-performance mapping
remains effective.

\subsection{Statistical Significance and Variance Analysis}
\label{app:stat}

To verify result stability, we repeat each experiment three times with different random seeds and report mean~$\pm$~standard deviation on three representative benchmarks in Table~\ref{tab:variance_id} and \ref{tab:variance_ood}.

\begin{table}[t!]
\centering
\caption{\textbf{In-distribution performance comparison} (mean~$\pm$~std, three runs).}
\label{tab:variance_id}
\small
\resizebox{1.0\linewidth}{!}{
\begin{tabular}{lccc}
\toprule
Method & AIME2025 & AMC & HumanEval \\
\midrule
RouterDC       & $22.2 \pm 1.9$ & $62.4 \pm 4.3$ & $80.5 \pm 1.6$ \\
MLPRouter      & $ 8.9 \pm 1.9$ & $45.0 \pm 2.5$ & $76.0 \pm 1.9$ \\
BertRouter     & $13.3 \pm 3.4$ & $44.2 \pm 1.4$ & $75.6 \pm 1.2$ \\
\rowcolor[RGB]{245, 238, 248}
\textbf{\textsc{Atlas}(Cluster)} 
               & $\mathbf{38.9 \pm 1.9}$ 
               & $\mathbf{82.5 \pm 2.5}$ 
               & $\mathbf{91.5 \pm 1.6}$ \\
\bottomrule
\end{tabular}}
\end{table}

\begin{table}[t!]
\centering
\caption{\textbf{Out-of-distribution performance comparison} (mean~$\pm$~std, three runs).}
\label{tab:variance_ood}
\small
\resizebox{1.0\linewidth}{!}{
\begin{tabular}{lccc}
\toprule
Method & AIME2025 & AMC & HumanEval \\
\midrule
RouterDC       & $ 2.2 \pm 1.9$ & $46.7 \pm 1.4$ & $79.1 \pm 0.9$ \\
MLPRouter      & $ 3.3 \pm 3.4$ & $32.5 \pm 2.5$ & $75.1 \pm 1.9$ \\
BertRouter     & $ 7.8 \pm 1.9$ & $39.2 \pm 1.4$ & $78.7 \pm 1.6$ \\
\rowcolor[RGB]{236,244,252}
\textsc{Atlas}(Cluster) 
               & $ 3.3 \pm 3.4$ & $47.5 \pm 2.5$ & $\mathbf{91.5 \pm 0.9}$ \\
\rowcolor[RGB]{245, 238, 248}
\textbf{\textsc{Atlas}(RL)} 
               & $\mathbf{33.3 \pm 3.4}$ 
               & $\mathbf{67.5 \pm 2.5}$ 
               & $85.4 \pm 0.6$ \\
\bottomrule
\end{tabular}}
\end{table}

Beyond variance, we apply the Wilcoxon signed-rank test~\cite{wilcoxon1992individual} pairing \textsc{Atlas} against \textsc{RouterDC} across all 10 benchmarks in Table~\ref{tab:main_results}. The in-domain test yields $p = 9.7 \times 10^{-4}$ ($p < 0.05$), rejecting the null hypothesis $H_0$ of no significant difference. The OOD setting yields equivalent significance ($p < 0.05$), validating both the robustness and generalization capability of our framework.

\section{Case Study}\label{sec:appendix_case_study}
\begin{figure*}[htbp!]
\centering
\includegraphics[width=0.98\textwidth]{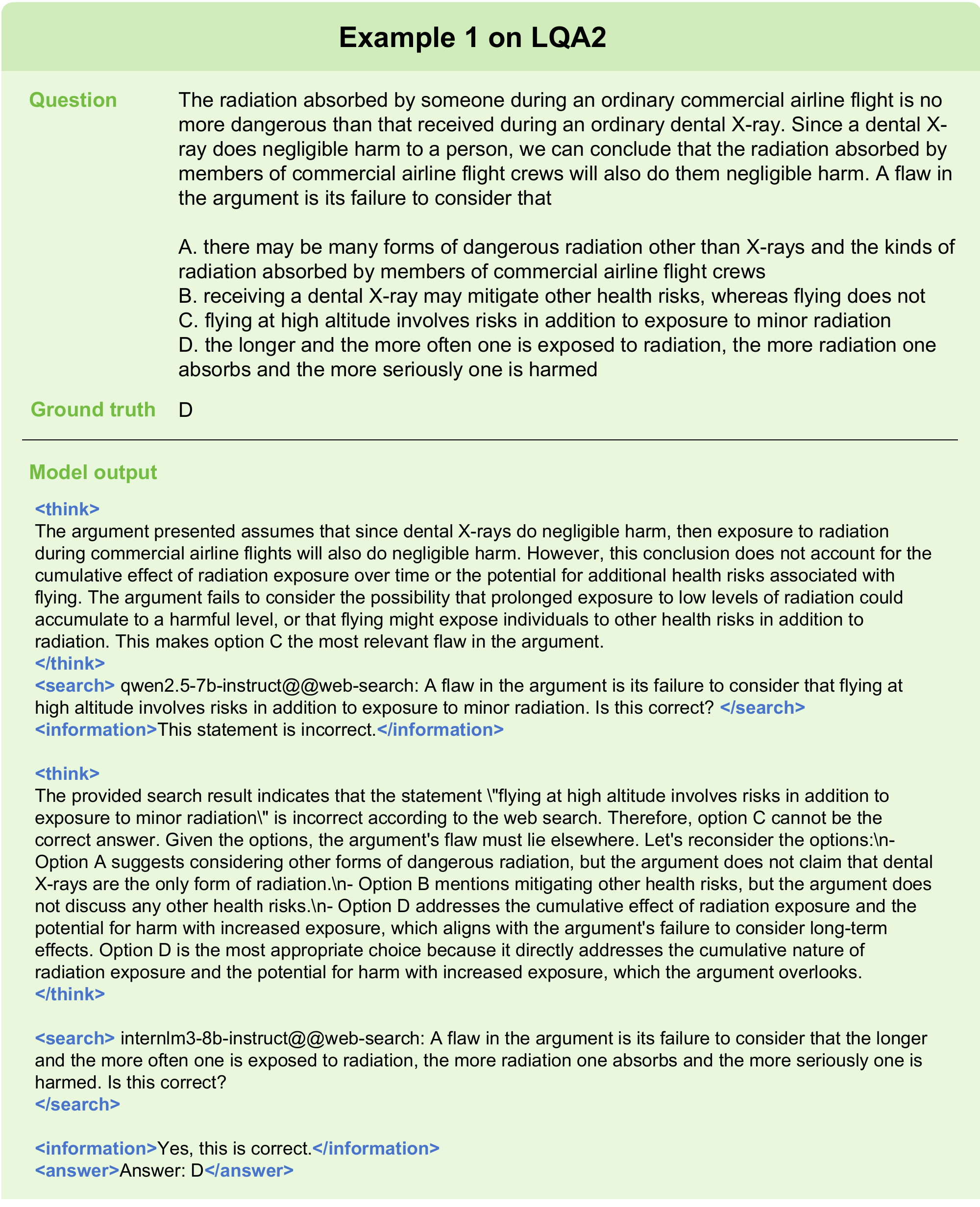}
\caption{Example 1 on the LQA2 dataset.}
\label{case_lqa2}
\end{figure*}

\begin{figure*}[htbp!]
\centering
\includegraphics[width=0.98\textwidth]{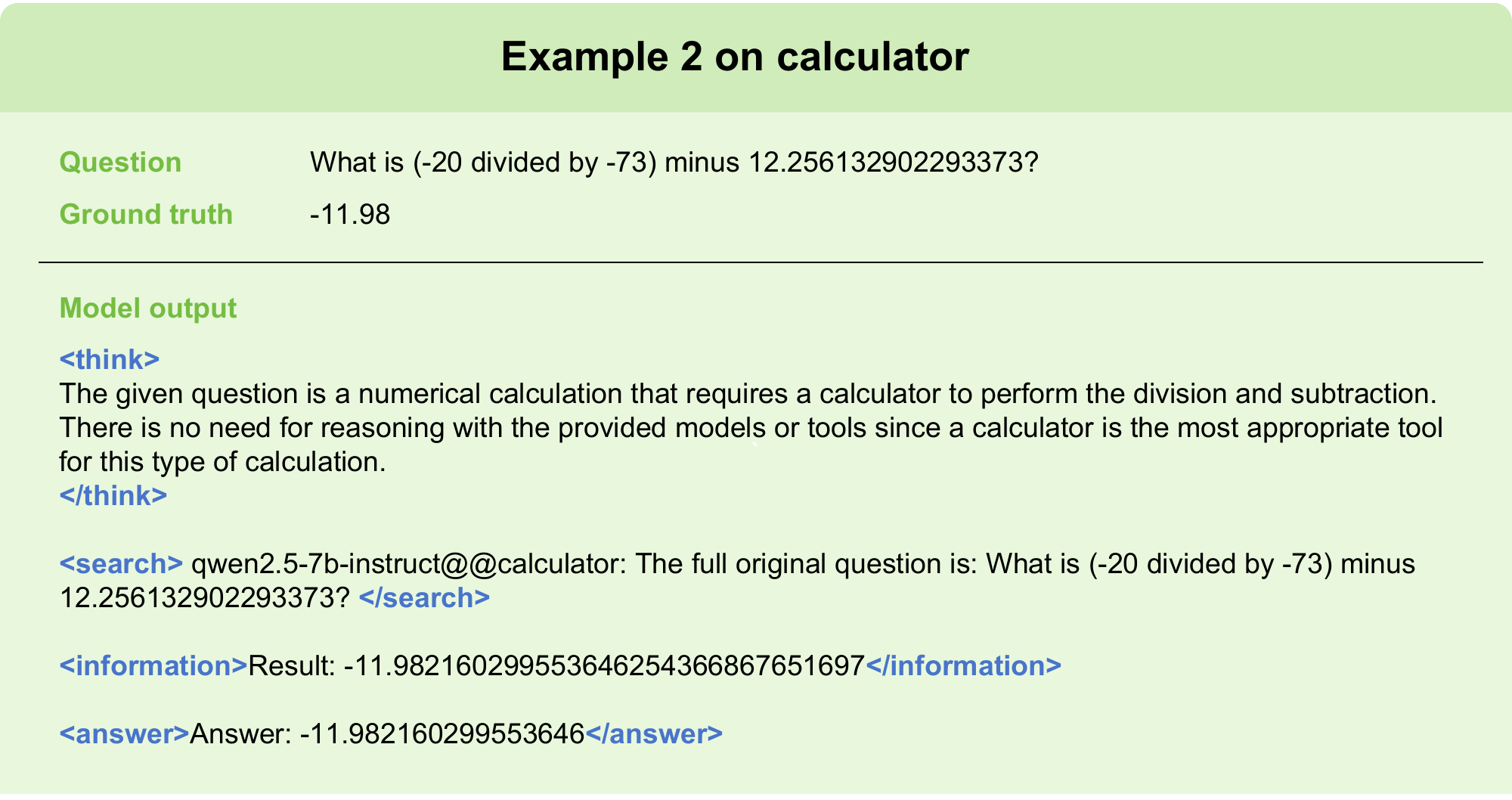}
\caption{Example 2 on the Calculator dataset.}
\label{case_calculator}
\end{figure*}

\begin{figure*}[htbp!]
\centering
\includegraphics[width=0.98\textwidth]{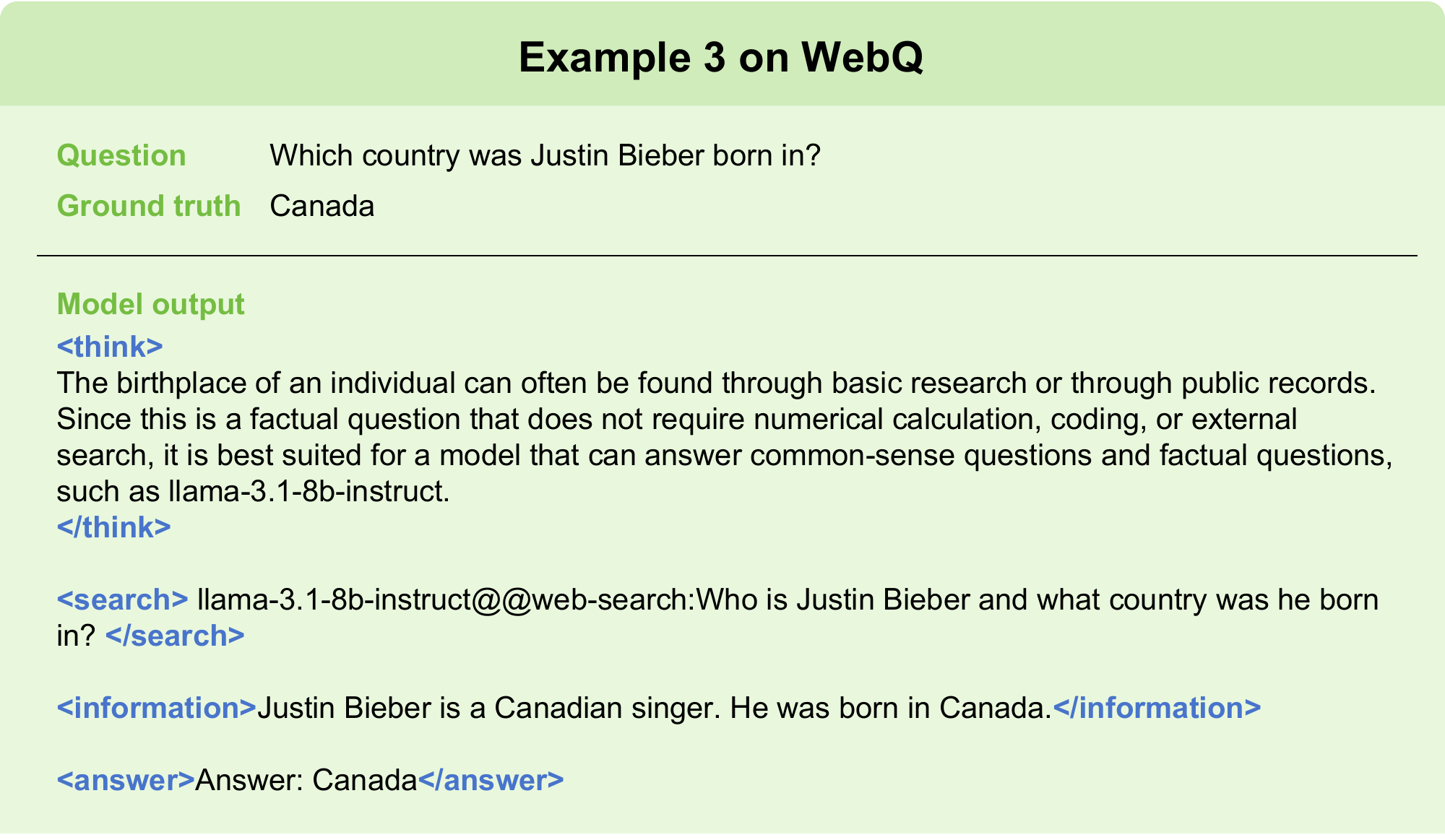}
\caption{Example 3 on the WebQ dataset.}
\label{case_webq}
\end{figure*}

\begin{figure*}[htbp!]
\centering
\includegraphics[width=0.98\textwidth]{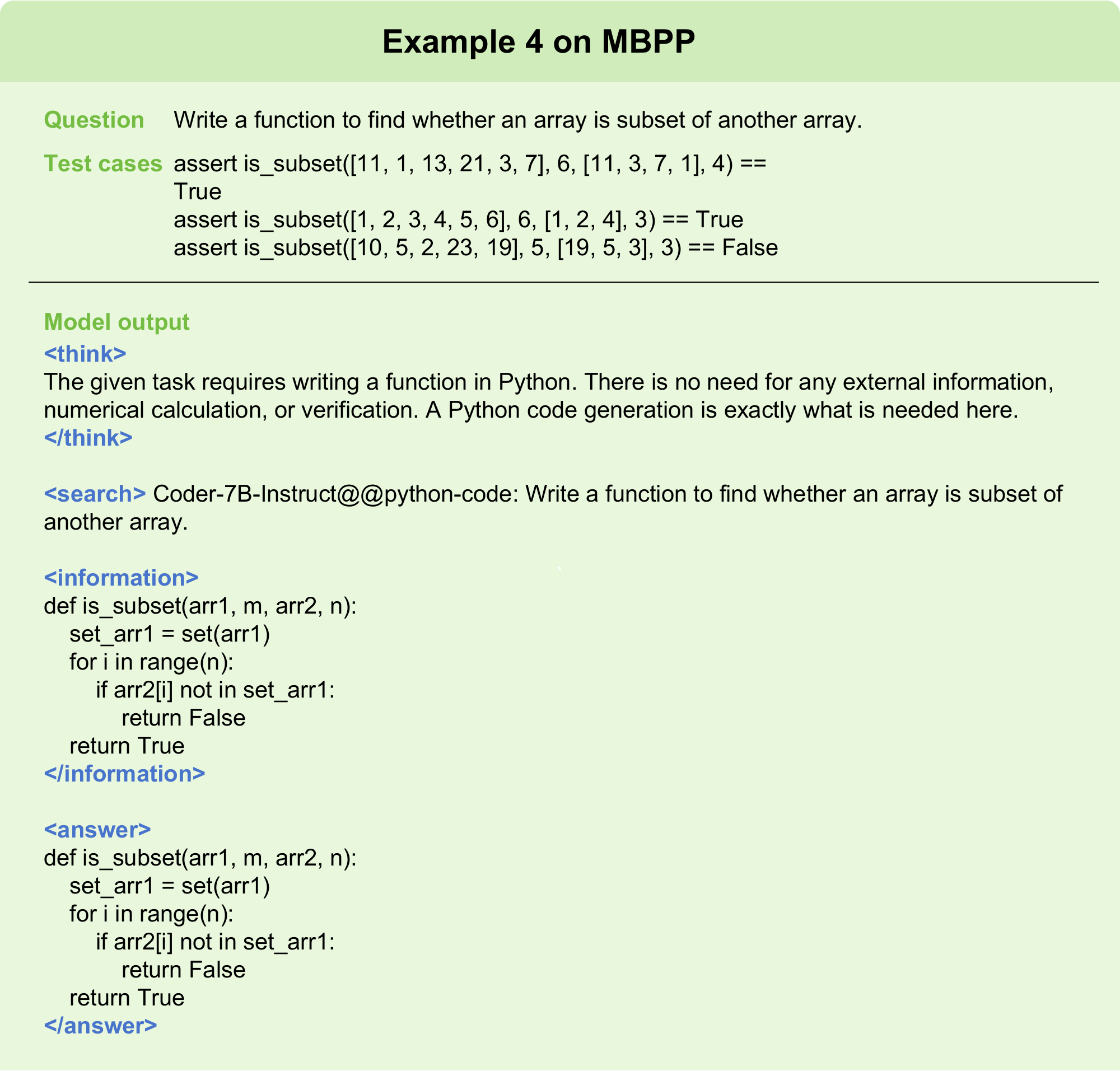}
\caption{Example 4 on the MBPP dataset.}
\label{case_mbpp}
\end{figure*}

\begin{figure*}[htbp!]
\centering
\includegraphics[width=0.98\textwidth]{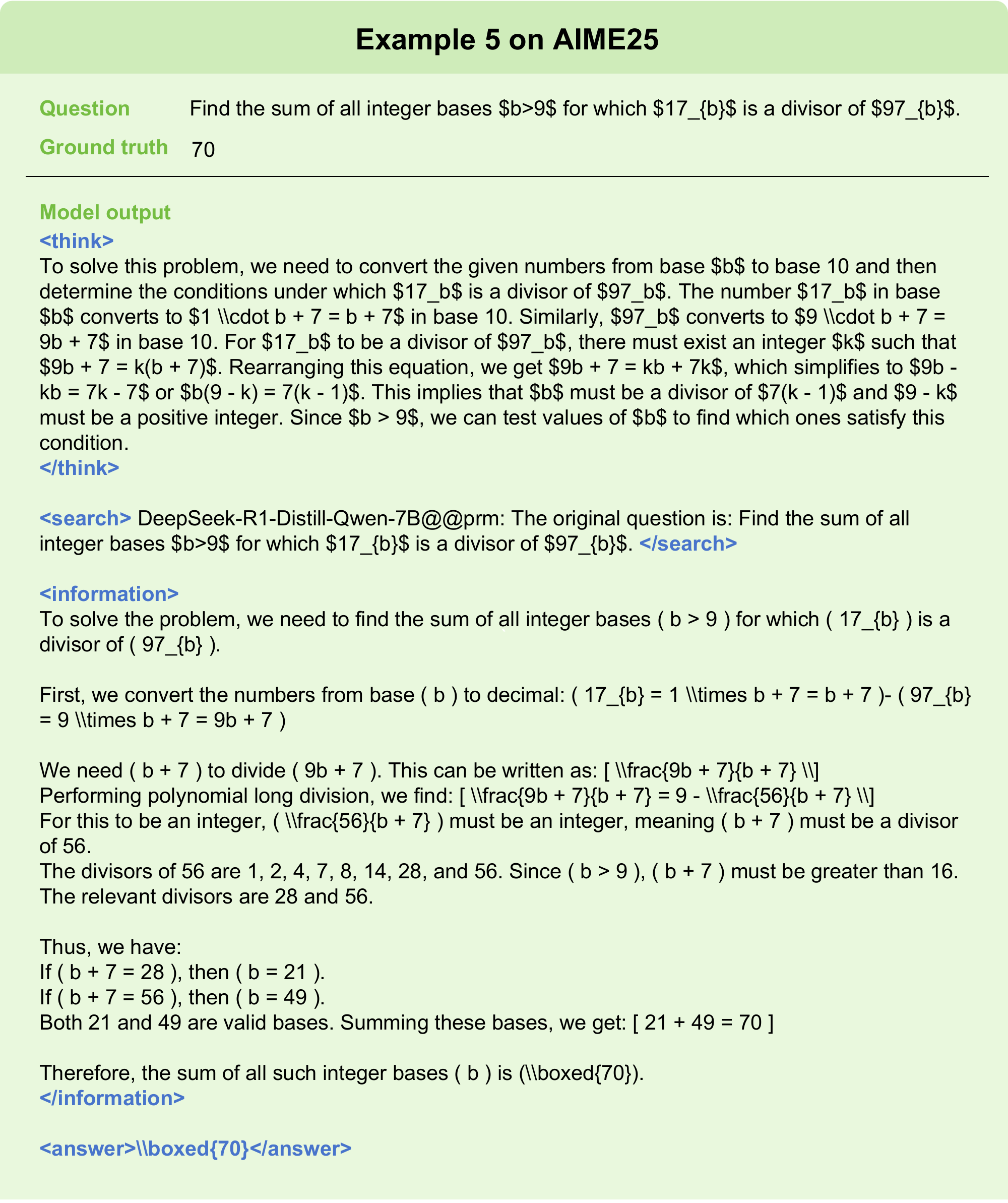}
\caption{Example 5 on the AIME dataset.}
\label{case_aime}
\end{figure*}

We provide some representative examples in Figures~\ref{case_lqa2}--\ref{case_aime} to illustrate how \textsc{Atlas} dynamically orchestrates model-tool combinations across diverse reasoning tasks.

\paragraph{Adaptive Multi-turn Reasoning.} Figure~\ref{case_lqa2} demonstrates \textsc{Atlas}'s capacity for self-correction through iterative exploration. When addressing a logical reasoning problem, the policy initially selects Qwen2.5-7B with web search to verify its hypothesis (option C), but upon receiving contradictory feedback, it re-evaluates the alternatives and routes to InternLM3-8B for a second verification. This multi-turn deliberation ultimately leads to the correct answer (option D), showcasing the framework's ability to recover from suboptimal initial decisions through adaptive re-routing.

\paragraph{Task-Aware Model-Tool Alignment and Selection.} Figures~\ref{case_calculator}--\ref{case_aime} highlight how \textsc{Atlas} aligns model-tool pairs with task-specific requirements. For arithmetic computation (Figure~\ref{case_calculator}), the policy directly invokes the calculator tool without unnecessary reasoning steps. For factual retrieval (Figure~\ref{case_webq}), it routes to Llama-3.1-8B with web search, recognizing the need for external knowledge. Code generation tasks (Figure~\ref{case_mbpp}) are delegated to the specialized Qwen2.5-Coder model with Python execution. For challenging mathematical problems (Figure~\ref{case_aime}), \textsc{Atlas} combines DeepSeek-7B with PRM for rigorous verification. These examples collectively demonstrate that \textsc{Atlas} has internalized meaningful associations between task categories and optimal model-tool configurations, rather than relying on rigid heuristics.

\section{Additional Discussion}
\subsection{Distinguishing ATLAS from Prior Routing and Tool Usage Methods}\label{subsec:novelty_discussion}
While \textsc{Atlas} employs established techniques such as semantic clustering for query representation and PPO for policy optimization, its contribution extends beyond the individual components to address a fundamental gap in existing literature: the joint optimization of heterogeneous model-tool combinations. Prior routing methods~\citep{chen2024routerdc,ong2024routellm,lu2024routing} focus exclusively on model selection, treating LLMs as isolated execution units without considering external tool augmentation. Conversely, tool usage frameworks~\citep{feng2025retool,wu2025tool} rely on fixed invocation logic that cannot dynamically adapt to different model capabilities. \textsc{Atlas} unifies these two paradigms by explicitly modeling the Cartesian product space $\mathcal{S} = \mathcal{M} \times \mathcal{T}$ and learning task-aware alignments within this joint space.

The technical novelty of \textsc{Atlas} manifests in three key aspects. \textbf{First}, the dual-path architecture strategically combines training-free cluster-based routing for exploiting domain-specific priors with RL-driven exploration for generalizing to unfamiliar tasks, achieving complementary strengths across distribution shifts (Table~\ref{tab:main_results}). \textbf{Second}, our composite reward structure ($\mathcal{R}_{\text{fmt}} + \gamma \mathcal{R}_{\text{out}} + \xi \mathcal{R}_{\text{sel}}$) decouples execution correctness from routing efficiency through the $\mathcal{R}_{\text{sel}}$ signal, enabling the policy to internalize transferable expertise distribution rather than memorizing task-specific mappings, which is evidenced by robust generalization to expanded model-tool pools without retraining (Section~\ref{subsec:dynamic_synergy}). \textbf{Third}, our controlled experiments ensure that all configurations share identical backbone models and evaluation protocols, with only routing mechanisms differing (Section~\ref{subsec:appendix_implementation_details}), thereby isolating the contribution of orchestration strategies from confounding factors such as model capacity or prompt engineering. The consistent performance gains across 15 benchmarks, including out-of-distribution settings (+13.1\% over baselines) and multi-modal tasks (+4.3\%), demonstrate that \textsc{Atlas} captures generalizable principles for adaptive model-tool coordination.

\subsection{Discussion on RL Reward Design}\label{subsec:reward_discussion}
Our composite reward function $r_\phi = \mathcal{R}_{\text{fmt}} + \gamma \mathcal{R}_{\text{out}} + \xi \mathcal{R}_{\text{sel}}$ balances structured execution, task correctness, and routing efficiency. Regarding potential concerns about $\mathcal{R}_{\text{out}}$ that require ground-truth labels, we note that test-time reinforcement learning remains effective in label-scarce scenarios through alternative supervision signals: majority voting across sampled trajectories has proven effective as pseudo-labeling~\citep{zuo2025ttrl}. Future extensions of \textsc{Atlas} could integrate such self-verification mechanisms to further reduce reliance on explicit supervision.

Regarding $\mathcal{R}_{\text{sel}}$, which penalizes suboptimal model selections based on offline evaluation (Equation~\ref{eq_sel}), a potential concern is whether this introduces evaluator bias or test-time information leakage. However, $\mathcal{R}_{\text{sel}}$ encodes domain priors from offline profiling, such as ``code tasks benefit from specialized models'' or ``retrieval tasks favor search-augmented models'', which practitioners naturally possess and use to initialize routing systems. Critically, it does not leak test-time information but rather provides consistent training targets to guide efficiency-aware exploration. The low weight $||\xi \mathcal{R}_{\text{sel}}||=0.15$ (vs. $\||\gamma \mathcal{R}_{\text{out}}||=1.0$ for $\mathcal{R}_{\text{out}}$) ensures routing efficiency serves as an auxiliary signal without overriding correctness. Our ablation (Figure~\ref{fig:reward_conv}-\ref{fig:entropy_loss}) shows that $\mathcal{R}_{\text{sel}}$ accelerates convergence and reduces entropy, while generalization to expanded model pools without prior annotations (Table~\ref{tab:extension_results}) shows that the policy learns transferable routing principles that aligns task characteristics with model capabilities, rather than memorizing specific mappings.

\subsection{When to Use Cluster-Based vs. RL-Based Routing}\label{subsec:routing_selection}
The choice between cluster-based and RL-based routing depends on data availability and generalization requirements. When domain-specific training data is accessible, such as historical query-answer pairs in enterprise QA systems, cluster-based routing offers a simple and efficient solution. It achieves strong in-domain performance (63.5\% average accuracy, Table~\ref{tab:main_results}) with zero training cost by leveraging semantic clustering and historical statistics, making it ideal for rapid deployment in well-defined domains. Conversely, when the reasoning engine must handle diverse, unfamiliar tasks where domain priors are unavailable, such as general-purpose assistants facing unpredictable queries, RL-based routing provides superior generalization. It learns transferable patterns of when to invoke tools or defer to specialized models, maintaining robust OOD performance (59.4\% vs. 49.2\% for cluster-based) at the cost of upfront training. In practice, practitioners can adopt a hybrid strategy: using cluster-based routing as the default for efficiency while reserving RL-based routing for critical queries or new domains, thereby balancing simplicity with adaptability.

\subsection{Future Work}\label{subsec:future_works}
Several promising directions remain for future exploration. First, the current web-search tool does not account for retrieval noise; integrating noise-aware retrieval strategies~\cite{li-etal-2025-survey,wu-etal-2025-pandoras} could further improve the robustness of information-augmented reasoning. Second, the RL routing policy could benefit from structured template guidance~\cite{yan2025learning,wu2025templaterl} to impose stronger inductive biases over multi-step trajectories, potentially accelerating convergence and improving sample efficiency. Third, evaluating \textsc{Atlas} in long-horizon, interactive settings such as \textsc{OdysseyArena}~\cite{xu2026odysseyarena,wu2026spark,lu2026skill0,yan2026tide} would provide a more rigorous assessment of its agentic capabilities beyond static benchmarks. On the efficiency side, speculative parallelism techniques~\cite{shen2026double,shen2026specbranch} offer a natural avenue to reduce the multi-step inference latency incurred by the RL-driven routing path. Finally, extending the framework to richer perceptual abilities---such as unified visual agents that tightly couple perception with reasoning~\cite{reddy2025orion,lu2025uis1,lu2026uir1,chen2025gui,gan2026android}---remains an important direction toward building truly general-purpose orchestration systems.

%% file: section/algorithm_training_free.tex
\begin{algorithm*}[t]
    \caption{Training-free Cluster-based Routing}
    \label{alg:cluster_routing_full}
    \begin{algorithmic}[1]
    \renewcommand{\algorithmicrequire}{\textbf{Input:}} 
    \renewcommand{\algorithmicensure}{\textbf{Output:}}
    
    \Require Test query $q_j$, cluster centroids $\{\mu_1, \dots, \mu_K\}$, historical performance statistics $\mathrm{Stats}[k][(m,t)]$, performance-cost trade-off parameter $\alpha$;
    \Ensure Optimal model-tool pair $(m^*, t^*)$ and the generated response $y_j$;
    
    \Statex \textit{\textcolor{mintleaf}{// Step 1: Query Representation}}
    \State $\mathbf{v}_j \gets \mathrm{Embed}(q_j)$ \Comment{Project query into the latent embedding manifold}
    
    \Statex \textit{\textcolor{mintleaf}{// Step 2: Semantic Clustering}}
    \State $k^* \gets \arg\min_{k \in \{1,\dots,K\}} \mathrm{dist}(\mathbf{v}_j, \mu_k)$ \Comment{Find the nearest semantic cluster centroid}
    
    \Statex \textit{\textcolor{mintleaf}{// Step 3: Dynamic Selection}}
    \If{$\mathrm{Stats}[k^*]$ is not empty}
        \For{each candidate pair $(m, t) \in \mathcal{M} \times \mathcal{T}$}
            \State $\mathcal{U}_{k^*}^{(m,t)} \gets (1-\alpha)\cdot\mathrm{Accuracy}(\mathrm{Stats}[k^*])-\alpha\cdot\mathrm{Cost}(\mathrm{Stats}[k^*])$
        \EndFor
        \State $(m^*, t^*) \gets \arg\max_{(m,t)} \mathcal{U}_{k^*}^{(m,t)}$ \Comment{Select the optimal combination}
    \Else
        \State $(m^*, t^*) \gets \mathrm{FallbackStrategy}(q_j)$ \Comment{Handle out-of-distribution queries}
    \EndIf
    \State $y_j \gets \mathrm{Execute}(m^*, t^*, q_j)$ \Comment{Invoke the selected model with the specific tool}    
    \State \Return $y_j$
    \end{algorithmic}
\end{algorithm*}

%% file: section/algorithm_rl.tex
\begin{algorithm*}[t]
    \caption{RL-driven Multi-step Routing}
    \label{alg:rl_routing_full}
    \begin{algorithmic}[1]
    \renewcommand{\algorithmicrequire}{\textbf{Input:}} 
    \renewcommand{\algorithmicensure}{\textbf{Output:}}
    
    \Require Query $q_j$, policy $\pi_\theta$, reference $\pi_{\text{ref}}$, pool $\mathcal{P}$, parameters $T_{\max}, \theta, \beta, \gamma, \xi$;
    \Ensure Response $y_j$ and trajectory $\tau$;
    
    \Statex \textit{\textcolor{mintleaf}{// Step 1: Initialization}}
    \State $\tau \gets \emptyset$, $C_0 \gets \emptyset$, $s_0 \gets \{q_j, C_0\}$
    
    \Statex \textit{\textcolor{mintleaf}{// Step 2: Multi-step Reasoning Loop}}
    \For{$t = 0$ to $T_{\max} - 1$}
        \State $a_t \sim \pi_\theta(\cdot \mid s_t, \mathcal{P})$ \Comment{Action $a_t \in \{\texttt{think}, \texttt{route}(m,t_\text{tool})\}$}
        \If{$a_t = \texttt{think}$}
            \State $o_t \gets \pi_\theta.\text{Reasoning}(s_t)$ \Comment{Internal reasoning}
        \Else
            \State $o_t \gets \mathrm{Execute}(m, t_\text{tool}, s_t)$ \Comment{Dynamic routing and tool invocation}
        \EndIf
        \State $C_{t+1} \gets C_t \cup \{a_t, o_t\}$, $s_{t+1} \gets \{q_j, C_{t+1}\}$
        \State $\tau \gets \tau \cup \{(s_t, a_t, o_t)\}$
        \State \textbf{if} $a_t$ contains \texttt{Final Answer} \textbf{then break}
    \EndFor
    
    \State $y_j \gets \mathrm{ParseAnswer}(\tau)$ \Comment{Answer extraction}
    
    \Statex \textit{\textcolor{mintleaf}{// Step 3: Policy Update (Training Mode)}}
    \If{\texttt{training\_mode}}
        \State $r_\phi(\tau) \gets \mathcal{R}_{\text{fmt}} + \gamma \mathcal{R}_{\text{out}} + \xi \mathcal{R}_{\text{sel}}$
        \State $\mathcal{L}_\theta \gets -\left[r_\phi(\tau) \cdot \log \pi_\theta(\tau) - \beta \cdot \log \frac{\pi_\theta(\tau)}{\pi_{\text{ref}}(\tau)}\right]$
        \State Update $\theta$ via PPO update rule: $\nabla_\theta \mathcal{L}_\theta$
    \EndIf
    
    \State \Return $(y_j, \tau)$
    \end{algorithmic}
\end{algorithm*}

%% file: section/x_datasets.tex
\begin{table*}[htbp!]
\caption{Detailed information on the datasets and test set sizes used in our experiments.}
\label{table:dataset-details}
\centering
\begin{adjustbox}{width=0.95\textwidth}
\begin{tabular}{llc}
\toprule
\textbf{Category} & \textbf{Dataset} & \textbf{\#Test Samples} \\
\midrule
\multirow{3}{*}{Mathematical Reasoning}
& AIME 2024~\citep{AIME2024} & 30 \\
& AIME 2025~\citep{AIME2025} & 30 \\
& AMC~\citep{lightman2023let} & 40 \\
\midrule
\multirow{2}{*}{Code Generation}
& HumanEval~\citep{chen2021evaluating} & 164 \\
& MBPP~\citep{austin2021program} & 974 \\
\midrule
Arithmetic Reasoning
& Calculator (Calc.)~\citep{wu2025tool} & 1000 \\
\midrule
\multirow{2}{*}{Commonsense Reasoning}
& Natural Questions (NQ)~\citep{kwiatkowski2019natural} & 1200 \\
& Web Question (WebQ)~\citep{webq} & 1000 \\
\midrule
Logical Reasoning
& LogiQA2~\citep{liu2023logiqa} & 1572 \\
\midrule
Scientific Reasoning
& GPQA~\citep{rein2024gpqa} & 448 \\
\midrule
\multirow{5}{*}{\makecell[c]{Multi-modal Perception\\and Reasoning}}
& ChartQA~\citep{masry2022chartqa} & 500 \\
& Geometry3K~\citep{lu2021inter} & 601 \\
& TallyQA~\citep{acharya2019tallyqa} & 498 \\
& CountBench~\citep{paiss2023teaching} & 491 \\
& TableVQA~\cite{kim2024tablevqa} & 500 \\
\bottomrule
\end{tabular}
\end{adjustbox}
\end{table*}

%% file: section/appendix_main_results.tex
\begin{table*}[t]
\small
\centering
\caption{\textbf{Extended performance comparison across diverse tasks and domains.} \textit{In-Distribution}: All datasets have training data available, so evaluation is in-distribution. \textit{Out-of-Distribution}: Models are trained only on Calc., NQ, and MBPP (in-distribution, marked as $\ddag$), then evaluated on all datasets (out-of-distribution for AIME24, AIME25, AMC, HumanEval, WebQ, LQA2, and GPQA). Zero-shot Router uses direct prompting without examples, while Few-shot Router uses prompting with examples. The best results are highlighted in \textbf{bold}.}
\label{tab:appendix_main_results}
\resizebox{1.0\linewidth}{!}{
\begin{tabular}{lccccccccccc}
\toprule
\multirow{2}{*}{\textbf{Method}} & \multicolumn{3}{c}{\textbf{Math Reasoning}} & \multicolumn{2}{c}{\textbf{Code}} & \textbf{Arith.} & \multicolumn{2}{c}{\textbf{Common.}} & \textbf{Logic} & \textbf{Sci.} & \multirow{2}{*}{\textbf{Avg.}} \\
\cmidrule(lr){2-4} \cmidrule(lr){5-6} \cmidrule(lr){7-7} \cmidrule(lr){8-9} \cmidrule(lr){10-10} \cmidrule(lr){11-11}
& AIME24 & AIME25 & AMC & Human. & MBPP$^\ddag$ & Calc.$^\ddag$ & NQ$^\ddag$ & WebQ & LQA2 & GPQA & \\
\midrule
\rowcolor{gray!8}\multicolumn{12}{c}{\textit{Closed-Source Models}}\\
Gemini2.5-Pro & 92.0 & 86.7 & 62.5 & 81.5 & 83.7 & 64.7 & 59.2 & 63.5 & 78.9 & 84.0 & 75.6 \\ 
Gemini2.5-Flash & 88.0 & 78.0 & 72.5 & 80.5 & 82.6 & 58.9 & 54.9 & 61.3 & 74.6 & 78.3 & 73.0 \\ 
GPT-5          & 93.3 & 94.6 & 97.5 & 93.4 & 98.4 & 82.9 & 59.3 & 61.5 & 83.8 & 85.7 & 85.0 \\ 
GPT-4.1        & 46.7 & 33.3 & 82.5 & 92.1 & 57.7 & 62.0 & 54.5 & 61.5 & 78.2 & 62.1 & 63.0 \\ 
GPT-4o         & 13.3 & 6.7 & 45.8 & 85.4 & 82.6 & 58.1 & 59.4 & 63.0 & 72.9 & 44.4 & 53.1 \\ 
\midrule
\midrule
\rowcolor{gray!8}\multicolumn{12}{c}{\textit{Training-free Baselines}} \\
ZS Router        & 13.3 & 6.7  & 32.5 & 53.0 & 64.2 & 55.7 & 29.2 & 39.2 & 45.3 & 24.6 & 36.4 \\
FS Router        & 23.3 & 13.3 & 40.0 & 68.9 & 64.7 & 47.2 & 27.3 & 35.8 & 40.8 & 25.9 & 38.7 \\
Random Router    & 6.7  & 3.3  & 15.0 & 37.8 & 52.6 & 40.2 & 25.3 & 32.1 & 49.2 & 30.6 & 29.3 \\
\midrule
\midrule
\rowcolor{gray!8}\multicolumn{12}{c}{\textit{In-Distribution Performance}} \\
RouterDC         & 40.0 & 23.3 & 62.5 & 80.5 & 77.7 & 74.9 & 41.2 & 47.6 & 47.2 & 39.1 & 53.4 \\
GraphRouter      & 30.0 & 16.7 & 50.0 & 78.7 & 75.0 & 72.3 & 37.5 & 49.6 & 45.8 & 37.1 & 49.3 \\
EmbedLLM         & 23.3 & 13.3 & 45.0 & 75.6 & 72.0 & 76.7 & 36.8 & 48.3 & 51.4 & 35.0 & 47.7 \\
MLPRouter        & 26.7 & 10.0 & 45.0 & 76.2 & 68.7 & 48.2 & 32.1 & 40.4 & 41.2 & 34.8 & 42.3 \\
BertRouter       & 30.0 & 13.3 & 45.0 & 75.4 & 72.1 & 77.1 & 38.9 & 50.4 & 47.1 & 36.6 & 48.6 \\
\rowcolor[RGB]{245, 238, 248}
\textbf{\textsc{Atlas} (cluster)} & \textbf{43.3} & \textbf{40.0} & \textbf{82.5} & \textbf{91.5} & \textbf{83.6} & \textbf{83.3} & \textbf{43.8} & \textbf{53.6} & \textbf{66.8} & \textbf{46.4} & \textbf{63.5} \\
\midrule
\rowcolor{gray!8}\multicolumn{12}{c}{\textit{Out-of-Distribution Performance}} \\
RouterDC         & 13.3 & 3.3  & 47.5 & 79.2 & 78.7 & 70.8 & 40.1 & 50.8 & 50.4 & 28.6 & 46.3 \\
GraphRouter      & 16.7 & 3.3 & 42.5 & 76.2 & 73.4 & 71.2 & 36.5 & 49.3 & 47.2 & 27.7 & 44.4 \\
EmbedLLM         & 13.3 & 3.3 & 45.0 & 79.9 & 73.0 & 79.1 & 41.4 & 50.2 & 51.5 & 31.7 & 46.8 \\
MLPRouter        & 13.3 & 3.3  & 32.5 & 75.0 & 67.7 & 54.6 & 37.3 & 43.7 & 38.9 & 26.8 & 39.3 \\
BertRouter       & 6.7  & 6.7  & 40.0 & 78.7 & 79.0 & 67.0 & 38.9 & 51.4 & 40.3 & 27.7 & 43.6 \\
\rowcolor[RGB]{236,244,252}
\textbf{\textsc{Atlas} (cluster)} & 13.3 & 3.3 & 47.5 & \textbf{91.5} & \textbf{83.6} & \textbf{83.3} & 43.8 & 51.4 & 45.6 & 29.0 & 49.2 \\
\rowcolor[RGB]{245, 238, 248}
\textbf{\textsc{Atlas} (RL)} & \textbf{43.3} & \textbf{33.3} & \textbf{67.5} & 85.4 & 81.8 & 81.6 & \textbf{44.1} & \textbf{52.2} & \textbf{62.7} & \textbf{42.0} & \textbf{59.4} \\
\bottomrule
\end{tabular}}
\end{table*}